\documentclass{l4dc2024}

\usepackage[top=1in,bottom=1in,left=1in,right=1in]{geometry}
\usepackage{amssymb}
\usepackage{amsmath}
\usepackage{mathtools}
\usepackage{xcolor}
\usepackage{tikz}
\usetikzlibrary{shapes.geometric, arrows}
\usepackage{adjustbox}
\usepackage{relsize}

\newcommand{\R}{\mathbb{R}}

\def\bx {\boldsymbol{x}}
\def\by {\boldsymbol{y}}

\def\bp {\boldsymbol{p}}

\def\bb {\boldsymbol{b}}
\def\ba {\boldsymbol{a}
}
\def\bz {\boldsymbol{z}}

\def\R {\mathbb{R}}

\def\Rn {\R^n}

\def\HHJ {H}
\def\JHJ {J}

\newcommand{\Sxx}{P} % x^2 term in S
\newcommand{\Sx}{\mathbf{q}} % x term in S
\newcommand{\Sc}{r} % const term in S

\def\lossfunc{\mathcal{L}}

\def\weight{\theta}
\def\weightvec{{\boldsymbol{\weight}}}

\def\regfunc{R}

\def\HJmom{\mathbf{p}}
\def\HJx{\mathbf{x}}
\def\HJt{t}
\def\HJu{\mathbf{u}}
\def\Hamiltonian{\HHJ}
\def\HJIC{\JHJ}

\def\LPoutspace{\R^m}
\def\weightspace{\Rn}
\def\HJstatespace{\weightspace}

\def\MLbasismat{\Phi}
\def\MLbasis{\phi}

\def\MLy{\by}

\tikzstyle{box} = [rectangle, rounded corners, minimum width=0cm, minimum height=1cm,text centered, draw=black, fill=none]
\tikzstyle{boxsmall} = [rectangle, rounded corners, minimum width=0cm, minimum height=0cm,text centered, draw=black, fill=none]
\tikzstyle{nobox} = [rectangle, rounded corners, minimum width=0cm, minimum height=1cm,text centered, draw=none, fill=none]
\tikzstyle{doublearrow} = [thick,<->,>=stealth]
\tikzstyle{dottedarrow} = [thick,<->,>=stealth,dotted]

\usepackage{longtable}% for long tables

\usepackage{booktabs}

\usepackage[load-configurations=version-1]{siunitx} % newer version
\newtheorem{rem}{Remark}
\newtheorem{prop}{Proposition}

\jmlrvolume{1}
\jmlryear{2024}
\jmlrworkshop{6th Annual Learning for Dynamics \& Control Conference}

\title[HJ PDEs for Continual SciML]{Leveraging Hamilton-Jacobi PDEs with time-dependent Hamiltonians for continual scientific machine learning}

 \author{\Name{Paula Chen\nametag{\footnotemark[1]\footnotetext[1]{These authors contributed equally to this work.}}}\Email{paula.x.chen.civ@us.navy.mil}\\
 \addr Naval Air Warfare Center Weapons Division, China Lake, CA 93555, USA
 \AND
 \Name{Tingwei Meng\nametag{\footnotemark[1]}}\Email{tingwei@math.ucla.edu}\\
 \addr Department of Mathematics, UCLA, Los Angeles, CA 90025, USA
 \AND 
 \Name{Zongren Zou\nametag{\footnotemark[1]}} \Email{zongren\_zou@brown.edu}\\
  \addr Division of Applied Mathematics, Brown University, Providence, RI 02912, USA 
  \AND
 \Name{J\'er\^ome Darbon} \Email{jerome\_darbon@brown.edu}\\
 \addr Division of Applied Mathematics, Brown University, Providence, RI 02912, USA
 \AND
 \Name{George Em Karniadakis} \Email{george\_karniadakis@brown.edu}\\
 \addr Division of Applied Mathematics and School of Engineering, Brown University, Providence, RI 02912, USA \\
 \addr Pacific Northwest National Laboratory, Richland, WA 99354, USA
 }

\begin{document}

\maketitle

\begin{abstract}
We address two major challenges in scientific machine learning (SciML): interpretability and computational efficiency. We increase the interpretability of certain learning processes by establishing a new theoretical connection between optimization problems arising from SciML  and a generalized Hopf formula, which represents the viscosity solution to a Hamilton-Jacobi partial differential equation (HJ PDE) with time-dependent Hamiltonian. Namely, we show that when we solve certain regularized learning problems with integral-type losses, we actually solve an optimal control problem and its associated HJ PDE with time-dependent Hamiltonian. This connection allows us to reinterpret incremental updates to learned models as the evolution of an associated HJ PDE and optimal control problem in time, where all of the previous information is intrinsically encoded in the solution to the HJ PDE. As a result, existing HJ PDE solvers and optimal control algorithms can be reused to design new efficient training approaches for SciML that naturally coincide with the continual learning framework, while avoiding catastrophic forgetting. As a first exploration of this connection, we consider the special case of linear regression and leverage our connection to develop a new Riccati-based methodology for solving these learning problems that is amenable to continual learning applications. We also provide some corresponding numerical examples that demonstrate the potential computational and memory advantages our Riccati-based approach can provide.
\end{abstract}
\begin{keywords}
Hamilton-Jacobi PDEs; generalized Hopf formula; continual learning; optimal control
\end{keywords}

\section{Introduction}
\label{sec:intro}
%\update{everywhere say optimal control first then Hj pde}
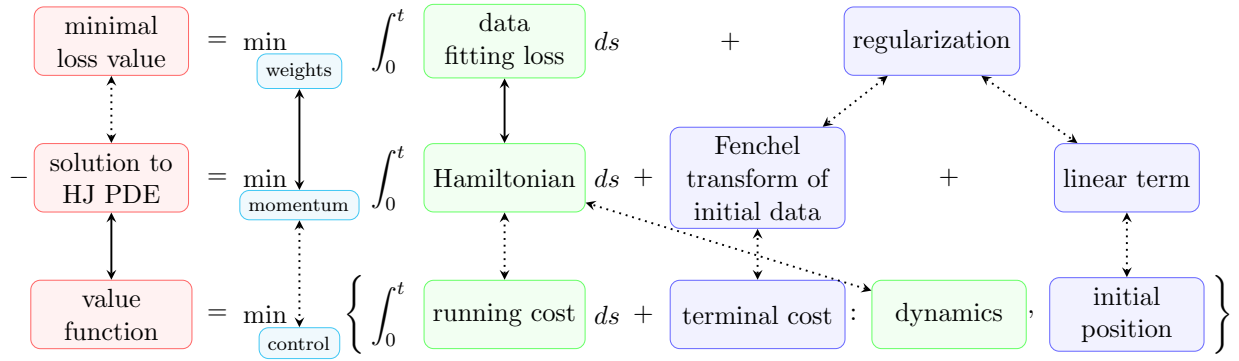
\begin{figure}[b!]
\begin{adjustbox}{width=\textwidth}
\begin{tikzpicture}[node distance=2cm]
    \node (min) [nobox, yshift=-0.2cm] {$\min$};
    \node (minarg) [boxsmall, right of=min, xshift=-1.5cm, yshift=-0.44cm, draw=cyan!60, fill=cyan!5] {$\text{}_\text{weights}$};
    \node (sum) [nobox, right of=min, xshift=-0.2cm] {$\mathlarger{\int}_0^t$};
    \node (loss) [box, right of=sum, xshift=-0.3cm, draw=green!60, fill=green!5, text width=2.1cm] {data fitting loss};
    \node (ds) [nobox, right of=loss, xshift=-0.5cm] {$ds$};
    \node (plus) [nobox, right of=loss, xshift=1.3cm] {$+$};
    \node (regLeft) [nobox, right of=loss, xshift=1.7cm] {};
    \node (regularization) [box, right of=regLeft, xshift=0.55cm, draw=blue!60, fill=blue!5] {regularization};
    \node (regRight) [nobox, right of=regularization, xshift=0.85cm] {};
    
    \node (sup) [nobox, below of=min, xshift=-0cm] {$\min$};
    \node (suparg) [boxsmall, right of=sup, xshift=-1.5cm, yshift=-0.4cm, draw=cyan!60, fill=cyan!5] {$\text{}_\text{momentum}$};
    \node (sum2) [nobox, below of=sum, xshift=0cm] {$\mathlarger{\int}_0^t$};
    \node (equals) [nobox, left of=sup, xshift=1.3cm] {$=$};
    \node (S) [box, left of=equals, xshift=0.45cm, text width=2cm, draw=red!60, fill=red!5] {solution to HJ PDE};
    \node (minus) [nobox, left of=S, xshift=0.65cm] {$-$};
    \node (Hamiltonian) [box, below of=loss, draw=green!60, fill=green!5, text width=2.1cm] {Hamiltonian};
    \node (HJds) [nobox, below of=ds] {$ds$};
    \node (plus2) [nobox, right of=Hamiltonian, xshift=0.03cm] {$+$};
    \node (IC) [box, below of=regLeft, text width=2.3cm, draw=blue!60, fill=blue!5] {Fenchel transform of initial data};
    \node (linear) [box, below of=regRight, draw=blue!60, fill=blue!5] {linear term};
    \node (plus3) [nobox, below of=regularization, xshift=0.25cm] {$+$};
    %\node (extraleft) [nobox, left of=minus, xshift=1.35cm, text width=0.8cm] {extra term};
    %\node (extraright) [box, below of=regRight, xshift=1.1cm, draw=blue!60, fill=blue!5, text width=0.8cm] {extra term};
    %\node (plusextra) [nobox, right of=linear, xshift=-0.83cm] {$+$};

    \node (minloss) [box, above of=S, text width=2cm, draw=red!60, fill=red!5] {minimal loss value};
    \node (LPequals) [nobox, above of=equals] {$=$};
    
    \node (OCmin) [nobox, below of=sup] {$\min$};
    \node (OCminarg) [boxsmall, right of=OCmin, xshift=-1.5cm, yshift=-0.42cm, draw=cyan!60, fill=cyan!5] {$\text{}_\text{control}$};
    \node (OCS) [box, below of=S, xshift=0cm, draw=red!60, fill=red!5, text width=2.1cm] {value function};
    \node (OCequals) [nobox, below of=equals, xshift=0cm] {$=$};
    \node (OCint) [nobox, below of=sum2] {$\mathlarger{\int}_0^t$};
    \node (OCHamiltonian) [box, below of=Hamiltonian, draw=green!60, fill=green!5] {running cost};
    \node (OCds) [nobox, below of=HJds] {$ds$};
    \node (OCplus) [nobox, below of=plus2, xshift=0cm] {$+$};
    \node (OCIC) [box, below of=IC, draw=blue!60, fill=blue!5, text width=2.3cm] {terminal cost};
    \node (OCdynam) [box, right of=OCIC, xshift=0.8cm, text width=2cm, draw=green!60, fill=green!5] {dynamics};
    \node (OCinitialposition) [box, below of=linear, text width=2cm, draw=blue!60, fill=blue!5] {initial position};
    \node (comma) [nobox, right of=OCdynam, xshift=-0.8cm] {,};
    \node (leftbracket) [nobox, left of=OCint, xshift=1.5cm] {$\Bigg\{$};
    \node (colon) [nobox, right of=OCIC, xshift=-0.6cm] {$:$};
    \node (rightbracket) [nobox, right of=OCinitialposition, xshift=-0.6cm] {$\Bigg\}$};
    
    \draw [doublearrow] (loss) -- (Hamiltonian);
    \draw [dottedarrow] (regularization) -- (IC);
    \draw [dottedarrow] (regularization) -- (linear);
    \draw [doublearrow] (minarg) -- (suparg);
    \draw [doublearrow] (OCS) -- (S);
    \draw [dottedarrow] (OCHamiltonian) -- (Hamiltonian);
    \draw [dottedarrow] (OCIC) -- (IC);
    \draw [dottedarrow] (OCdynam) -- (Hamiltonian);
    \draw [dottedarrow] (OCinitialposition) -- (linear);
    \draw [dottedarrow] (S) -- (minloss);
    \draw [dottedarrow] (suparg) -- (OCminarg);
\end{tikzpicture}
\end{adjustbox}

    \caption{(See Section~\ref{sec:theory}) Illustration of a connection between a regularized learning problem with integral-type loss (\textbf{top}), the generalized Hopf formula for HJ PDEs with time-dependent Hamiltonians (\textbf{middle}), and the corresponding optimal control problem (\textbf{bottom}). The colors indicate the associated quantities between each problem. For example, the optimal weights in the learning problem are equivalent to the momentum in the HJ PDE, which is related to the control in the optimal control problem (\textcolor{cyan}{cyan}). This color scheme is reused in the subsequent illustrations of our connection. The solid-line arrows denote direct equivalences. The dotted arrows represent additional mathematical relations.}
    \label{fig:intro_connection_in_words}
\end{figure}

Scientific machine learning (SciML) encompasses a wide range of powerful, data-driven techniques renowned for their ability to solve complicated problems that more traditional numerical methods cannot. Despite these successes, %scientific machine learning still faces two main challenges: limited interpretability and computational efficiency.
%machine learning algorithms are often treated as black boxes as 
developing the theoretical foundations of SciML is still an active area of research (e.g., see \cite{carvalho2019machine}). In this work, we increase the interpretability of certain learning processes by establishing a new theoretical connection between certain optimization problems arising from SciML  and a generalized Hopf formula, which represents the viscosity solution to a Hamilton-Jacobi partial differential equation (HJ PDE) with time-dependent Hamiltonian (Section~\ref{sec:theory}). 
HJ PDEs have been shown to have deep connections with many scientific disciplines, including but not limited to optimal control 
(\cite{Bardi1997Optimal}) and differential games (\cite{evans1984differentialgames}). Here, we leverage our new theoretical connection as well as the established connection between HJ PDEs and optimal control to show that when we solve certain regularized learning problems, we actually solve an optimal control problem and its associated HJ PDE with time-dependent Hamiltonian. In doing so, we can reuse existing efficient HJ PDE solvers and optimal control algorithms to develop new training approaches for SciML.
We also build upon our previous work in~\cite{chen2023leveraging}, which, to our knowledge, is the first to establish a connection of this kind between SciML and HJ PDEs. In~\cite{chen2023leveraging}, we instead connect regularized learning problems with the multi-time Hopf formula, which represents the solution to certain multi-time HJ PDEs (\cite{rochet1985multitimeHJ, lions1986hopf}). By discretizing the generalized Hopf formula (e.g., by discretizing the integral in the middle row of  Figure~\ref{fig:intro_connection_in_words}), we can regard the work here as a generalization of~\cite{chen2023leveraging} to the infinite-time case. 

By instead considering HJ PDEs with time-dependent Hamiltonians, we develop a new theoretical framework that is more closely aligned with continual learning (\cite{parisi2019continual, van2019three}), which in turn yields potential computational and memory advantages, especially in big data regimes. Under the continual learning framework, data is accessed in a stream and learned models are updated incrementally as new data becomes available. In many continual learning scenarios, data is also assumed to become inaccessible after it is incorporated into the learned model, which often leads to catastrophic forgetting (\cite{kirkpatrick2017overcoming, parisi2019continual}) or, in other words, the abrupt degradation in the performance of learned models on previous tasks upon training on new tasks.
However, this lack of dependence on historical data often also facilitates the development of more computationally efficient learning algorithms by requiring that only a small amount of data can be processed and stored at a time.
Using our theoretical connection, we reinterpret incrementally updating learned models as evolving an associated HJ PDE and optimal control problem in time, where all of the information from previous data points are inherently encoded in the solution to the HJ PDE. As such, our new interpretation has the potential to yield new efficient continual learning approaches that naturally avoid catastrophic forgetting. 

As a first exploration of this connection, we consider the special case of regularized linear regression problems with integral-type data fitting losses, which we show are connected to linear quadratic regulator (LQR) problems with time-dependent costs and dynamics (\cite{anderson2007optimal, tedrake2023underactuated}) (Section~\ref{sec:method}). We then leverage this connection to develop a new Riccati-based methodology that coincides with the continual learning framework, while inherently avoiding catastrophic forgetting. Finally, we demonstrate how this methodology can be applied to some SciML applications of interest (Section~\ref{sec:examples}) and discuss some possible future directions (Section~\ref{sec:conclusion}).

\section{Connection between the generalized Hopf formula and learning problems}\label{sec:theory}
In this section, %we establish a new theoretical connection between regularized learning problems with integral-type losses, the generalized Hopf formula for HJ PDEs with time-dependent Hamiltonians, and optimal control problems. 
we begin by introducing the generalized Hopf formula and reviewing the well-established connections between HJ PDEs and optimal control. Then, we introduce the learning problems of interest and formulate our new theoretical connection.

\subsection{Generalized Hopf formula for HJ PDEs with time-dependent Hamiltonians}
Consider the following HJ PDE with time-dependent Hamiltonian:
\begin{equation}\label{eq:HJPDE_timedependent}
    \begin{dcases}
        \frac{\partial S(\HJx, \HJt)}{\partial \HJt} + \Hamiltonian(\HJt, \nabla_\HJx S(\HJx, \HJt)) = 0 & \HJx\in\Rn, \HJt > 0, \\
        S(\HJx, 0) = \HJIC(\HJx) & \HJx\in\Rn,
    \end{dcases}
\end{equation}
where $\Hamiltonian:(0,\infty)\times\Rn\to \R$ is the Hamiltonian and $J:\Rn\to\R$ is the initial condition. Assume that $\Hamiltonian, \HJIC$ are continuous, $\HJIC$ is convex, and $\Hamiltonian(\HJt,\HJmom)$ is convex in $\HJmom$. Then, using the same proof as that for \cite[Proposition 1]{lions1986hopf}, it can be shown that the viscosity solution to this HJ PDE~\eqref{eq:HJPDE_timedependent} can be represented by the following generalized Hopf formula:
\begin{equation}\label{eq:generalizedHopf}
    S(\HJx, \HJt) = \sup_{\HJmom\in\Rn}\left\{\langle\HJx,\HJmom\rangle - \int_0^\HJt \Hamiltonian(s,\HJmom)ds - \HJIC^*(\HJmom)\right\} = -\inf_{\HJmom\in\Rn}\left\{\int_0^\HJt \Hamiltonian(s,\HJmom)ds + \HJIC^*(\HJmom) - \langle\HJx,\HJmom\rangle\right\},
\end{equation}
where $f^*$ denotes the Fenchel-Legendre transform of the function $f$; i.e., $f^*(\HJmom) = \sup_{\HJx\in\Rn} \{\langle \HJx, \HJmom\rangle - f(\HJx)\}$. Note that, as stated in a remark in~\cite{lions1986hopf}, this Hopf representation formula does not hold in general. However, under our convexity assumptions, the semigroup commutation still holds and hence, we can follow the same proof as that for \cite[Proposition 1]{lions1986hopf}. More details are provided in the proof of Proposition~\ref{prop:genHopf} below.
Also note that if the integral in~\eqref{eq:generalizedHopf} is discretized in time, then we recover a multi-time Hopf formula (e.g., see \cite{rochet1985multitimeHJ, lions1986hopf, chen2023leveraging, darbon2019decomposition}), and hence,~\eqref{eq:HJPDE_timedependent} can also be considered as a multi-time HJ PDE with infinitely many times. 

It is well-known that the value function of the following optimal control problem also satisfies~\eqref{eq:HJPDE_timedependent}:
\begin{equation}\label{eq:opt_ctrl}
    S(\HJx,\HJt) = \min_{\HJu(\cdot)}\left\{\int_0^\HJt L(s, \HJu(s))ds + \HJIC(\bx(\HJt)): \dot\bx(s) = f(s, \HJu(s)) \forall s\in(0,\HJt], \bx(0) = \HJx\right\},
\end{equation}
where $\HJx$ is the initial position, $\HJt$ is the time horizon, $\HJu:[0,\infty)\to\LPoutspace$ is the control, $\bx:[0,\infty)\to \Rn$ is the trajectory, the running cost $L$ and the source term $f$ of the dynamics are related to the Hamiltonian $\Hamiltonian$ by $\Hamiltonian(s,\HJmom) = \sup_{\HJu\in\R^m} \{\langle -f(s,\HJu), \HJmom\rangle - L(s,\HJu)\}$, and $\HJIC$ is now the terminal cost.

\begin{prop}\label{prop:genHopf}
    If $\Hamiltonian:(0,\infty)\times\Rn\to\R$ and $\HJIC:\Rn\to\R$ are continuous, $\HJIC$ is convex, and $\Hamiltonian(\HJt,\HJmom)$ is convex in $\HJmom$ then the generalized Hopf formula~\eqref{eq:generalizedHopf} is the viscosity solution to the time-dependent HJ PDE~\eqref{eq:HJPDE_timedependent}.
\end{prop}

\begin{proof}
    We follow the proof for \cite[Proposition 1]{lions1986hopf}. To do so, we will show that~\eqref{eq:generalizedHopf} is both a viscosity subsolution and viscosity supersolution of~\eqref{eq:HJPDE_timedependent}, where we use the characterization of viscosity (sub-/super-) solutions using generalized gradients \cite[Proposition I.18]{crandall1983viscosity}. Namely, $S$ is viscosity subsolution (supersolution, resp.) of~\eqref{eq:HJPDE_timedependent} if and only if for all $(\HJx_0, t_0)\in\Rn\times[0,\infty)$ and for every $(\bp,q)$ in the superdifferential (subdifferential, resp.) of $S$ at $(\HJx_0, t_0)$, $q + H(t_0, \bp) \geq 0$ ($\leq 0$, resp.). $S$ is a viscosity solution if it is both a viscosity subsolution and viscosity supersolution. 
    
    First, we will show that~\eqref{eq:generalizedHopf} is a viscosity subsolution of~\eqref{eq:HJPDE_timedependent}. Note that the generalized Hopf formula~\eqref{eq:generalizedHopf} can equivalently be written as 
    \begin{equation}
        S(\HJx, \HJt) = \left(\HJIC^* + \int_0^\HJt \Hamiltonian(s,\cdot)ds\right)^*(\HJx).
    \end{equation}
    From this formulation, it is clear that~\eqref{eq:generalizedHopf} is a convex function satisfying~\eqref{eq:HJPDE_timedependent} at every point of differentiability. Since all convex functions are subdifferentiable, if $S$ is superdifferentiable at $(\HJx_0, t_0)$, then it is also differentiable at that point and hence, $q + H(t_0, \bp) = \frac{\partial S(\HJx_0,t_0)}{\partial t} + H(t_0,\nabla_\HJx S(\HJx_0,t_0)) = 0$ for all $(\bp,q)$ in the superdifferential of $S$ at $(\HJx_0, t_0)$. In other words, $S$ is a viscosity subsolution.

    Next, we will show that~\eqref{eq:generalizedHopf} is a viscosity supersolution of~\eqref{eq:HJPDE_timedependent}. Denote by $\Bar{S}_H(t_1,t_2)$, the operator $\Bar{S}_H(t_1,t_2)f = (f^* + \int_{t_1}^{t_2} H(\lambda, \cdot)d\lambda)^*$. Then, for all $s\in [0, t_0]$, we have that 
    \begin{equation} \label{eq:prop1_semigroup_identity}
        \begin{aligned}
            \left[\Bar{S}_H(t_0-s,t_0) S(\cdot, t_0-s)\right](\HJx_0) & = \left(S^*(\cdot, t_0-s) + \int_{t_0-s}^{t_0} H(\lambda, \cdot)d\lambda\right)^*(\HJx_0) \\
            & = \left( \left(\HJIC^* + \int_0^{\HJt_0-s} \Hamiltonian(\lambda,\cdot)d\lambda\right)^{**} + \int_{t_0-s}^{t_0} H(\lambda, \cdot)d\lambda\right)^*(\HJx_0) \\
            & = \left( \HJIC^* + \int_0^{\HJt_0-s} \Hamiltonian(\lambda,\cdot)d\lambda + \int_{t_0-s}^{t_0} H(\lambda, \cdot)d\lambda\right)^*(\HJx_0) \\
            & = \left( \HJIC^* + \int_0^{\HJt_0} \Hamiltonian(\lambda,\cdot)d\lambda\right)^*(\HJx_0) \\
            & = S(\HJx_0, t_0),
        \end{aligned}
    \end{equation}
    where the third equality follows from $f^{**} = f$ when $f$ is convex and lower semicontinuous. Let $(\bp,q)$ be in the subdifferential of $S$ at $(\HJx_0,t_0)$. Then, by the definition of subdifferentials, we have that for all $(\HJx,t)\in\Rn\times[0,\infty)$,
    $$S(\HJx,t) \geq S(\HJx_0,t_0) + \langle \bp, \HJx-\HJx_0\rangle + q(t-t_0).$$
    In particular, take $t = t_0 - s$ and apply $\Bar{S}_H(t_0-s,t_0)$ to both sides (note that $\Bar{S}_H(t_1,t_2)$ is order-preserving for all $t_1,t_2$ since the Fenchel-Legendre transform is order-reversing: $f\leq g \implies f^*\geq g^* \implies f^* + \int H(\lambda,\cdot)d\lambda \geq g^*  + \int H(\lambda,\cdot)d\lambda \implies  (f^* + \int H(\lambda,\cdot)d\lambda)^* \leq (g^*  + \int H(\lambda,\cdot)d\lambda)^*$) to get:
    \begin{equation}\label{eq:prop1_inequality}
    \begin{aligned}
        S(\HJx_0, t_0) & = \left[\Bar{S}_H(t_0-s,t_0)S(\cdot,t_0-s)\right](\HJx_0) \\
        & \geq  \left[\Bar{S}_H(t_0-s,t_0)\phi\right](\HJx_0) = \sup_{\bz\in\Rn}\left\{ \langle \bz,\HJx_0\rangle - \int_{t_0-s}^{t_0} H(\lambda,\bz)d\lambda - \phi^*(\bz)\right\},
    \end{aligned}
    \end{equation}
    where the first equality follows from~\eqref{eq:prop1_semigroup_identity}, we define $\phi(\HJx) = S(\HJx_0,t_0) + \langle \bp, \HJx-\HJx_0\rangle -sq$, and the second equality follows from the definitions of $\Bar{S}_H(t_0-s,t_0)$ and the Fenchel-Legendre transform.
    Now note that 
    \begin{equation}
    \begin{aligned}
    \phi^*(\bz) & = \sup_{\HJx\in\Rn} \{\langle \HJx,\bz\rangle - S(\HJx_0,t_0) - \langle \bp, \HJx-\HJx_0\rangle + sq\} \\
        & = \begin{dcases}
            sq - S(\HJx_0,t_0) + \langle \bp, \HJx_0\rangle & \bz = \bp \\
            +\infty & \bz\neq \bp.
        \end{dcases}
    \end{aligned}
    \end{equation}
    Combining this fact with~\eqref{eq:prop1_inequality} gives us that
    \begin{equation}
        S(\HJx_0,t_0) \geq \langle \bp,\HJx_0\rangle - \int_{t_0-s}^{t_0} H(\lambda,\bp)d\lambda - \phi^*(\bp) = S(\HJx_0,t_0) - sq - \int_{t_0-s}^{t_0} H(\lambda,\bp)d\lambda.
    \end{equation}
    Rearranging terms gives us that
    $$q + \frac{1}{s}\int_{t_0-s}^{t_0} H(\lambda,\bp)d\lambda \geq 0.$$
    Letting $s\to 0$, we have that $q + H(t_0,\bp) \geq 0$, or in other words, $S$ is a viscosity supersolution. 
\end{proof}

\subsection{Connection to learning problems}\label{sec:general_connection}
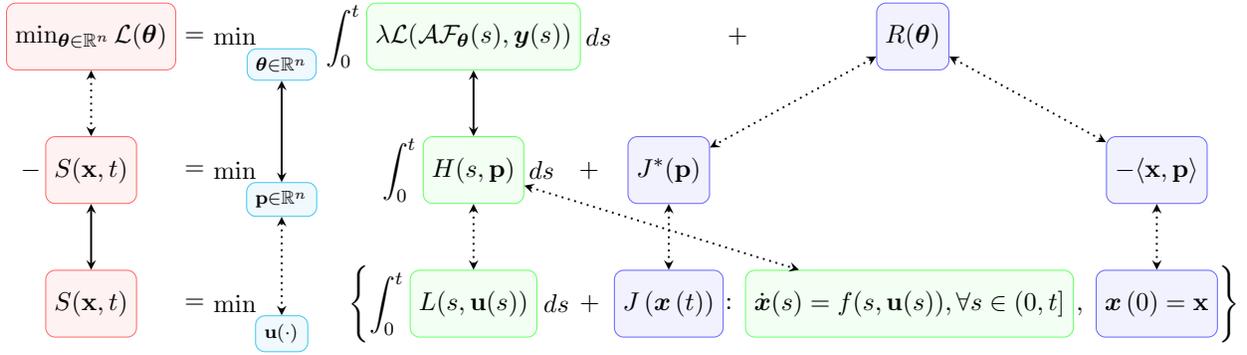
\begin{figure}[htbp]

    \centering   
    \begin{adjustbox}{width=\textwidth}
\begin{tikzpicture}[node distance=2cm]
    \node (min) [nobox, yshift=-0.2cm] {$\min$};
    \node (minarg) [boxsmall, right of=min, xshift=-1.3cm, yshift=-0.42cm, draw=cyan!60, fill=cyan!5] {$\text{}_{\weightvec\in\weightspace}$};
    \node (sum) [nobox, right of=min, xshift=-0.38cm] {$\mathlarger{\int}_0^t$};
    \node (loss) [box, right of=sum, xshift=-0.05cm, draw=green!60, fill=green!5] {$\lambda\lossfunc(\mathcal{A}\mathcal{F}_\weightvec(s), \by(s))$};
    \node(ds) [nobox, right of=loss, xshift=-0.13cm] {$ds$};
    \node (plus) [nobox, right of=loss, xshift=1.95cm] {$+$};
    \node (regLeft) [nobox, right of=loss, xshift=0.92cm] {};
    \node (regularization) [box, right of=regLeft, xshift=1.65cm, draw=blue!60, fill=blue!5] {$\regfunc(\weightvec)$};
    \node (regRight) [nobox, right of=regularization, xshift=1.65cm] {};
    
    \node (sup) [nobox, below of=min, xshift=-0cm] {$\min$};
    \node (suparg) [boxsmall, right of=sup, xshift=-1.3cm, yshift=-0.44cm, draw=cyan!60, fill=cyan!5] {$\text{}_{\HJmom\in\HJstatespace}$};
    \node (sum2) [nobox, below of=sum,xshift=0.84cm]{$\mathlarger{\int}_0^t$};
    \node (equal) [nobox, left of=sup, xshift=1.4cm] {$=$};
    \node (S) [box, left of=equal, xshift=0.45cm, draw=red!60, fill=red!5] {$S(\HJx, \HJt)$};
    \node (minus) [nobox, left of=S, xshift=1.1cm] {$-$};
    \node (Hamiltonian) [box, below of=loss, draw=green!60, fill=green!5] {$\Hamiltonian(s,\HJmom)$};
    \node(HJds) [nobox, below of=ds, xshift=-0.85cm] {$ds$};
    \node (plus2) [nobox, right of=Hamiltonian, xshift=-0.27cm] {$+$};
    \node (IC) [box, below of=regLeft, draw=blue!60, fill=blue!5] {$\HJIC^*(\HJmom)$};
    \node (linear) [box, below of=regRight, draw=blue!60, fill=blue!5] {$-\langle \HJx, \HJmom\rangle$};

    \node (OCmin) [nobox, below of=sup, xshift=0cm] {$\min$};
    \node (OCminarg) [boxsmall, below of=suparg, yshift=-0.01cm, draw=cyan!60, fill=cyan!5] {$\text{}_{\HJu(\cdot)}$};
    \node (OCS) [box, below of=S, xshift=0cm, draw=red!60, fill=red!5] {$S(\HJx, \HJt)$};
    \node (leftbracket) [nobox, right of=OCmin, xshift=-0.15cm] {$\Bigg\{$};
    \node (OCequal) [nobox, below of=equal] {$=$};
    \node (OCint) [nobox, below of=sum2, xshift=-0.19cm] {$\mathlarger{\int}_0^t$};
    \node (OCHamiltonian) [box, below of=Hamiltonian, draw=green!60, fill=green!5] {$L(s,\HJu(s)) $};
    \node (OCds) [nobox, right of=OCHamiltonian, xshift=-0.76cm] {$ds$};
    \node (OCplus) [nobox, right of=OCds, xshift=-1.5cm] {$+$};
    \node (OCIC) [box, below of=IC, draw=blue!60, fill=blue!5] {$\HJIC\left(\bx\left(\HJt\right)\right)$};
    \node (colon) [nobox, right of=OCIC, xshift=-1.1cm] {:};
    \node(OCdynamics) [box, right of=OCIC, xshift=1.6cm, draw=green!60, fill=green!5] {$\dot\bx(s) = f(s,\HJu(s)), \forall s \in \left(0, \HJt\right]$};
    \node(comma) [nobox, right of=OCdynamics, xshift=0.55cm, yshift=-0.11cm] {,};
    \node (terminalposition) [box, below of=linear, draw=blue!60, fill=blue!5] {$\bx\left(0\right) = \HJx$};
    \node (rightbracket) [nobox, right of=terminalposition, xshift=-0.92cm] {$\Bigg\}$};

    \node (LPequal) [nobox, above of=equal] {$=$};
    \node (minloss) [box, above of=S, draw=red!60, fill=red!5] {$\min_{\weightvec\in\weightspace}\mathcal{L}(\weightvec)$};
    
    \draw [doublearrow] (loss) -- (Hamiltonian);
    \draw [dottedarrow] (regularization) -- (IC);
    \draw [dottedarrow] (regularization) -- (linear);
    \draw [doublearrow] (minarg) -- (suparg);
    \draw [doublearrow] (S) -- (OCS);
    \draw [dottedarrow] (IC) -- (OCIC);
    \draw [dottedarrow] (Hamiltonian) -- (OCHamiltonian);
    \draw [dottedarrow] (Hamiltonian) -- (OCdynamics);
    \draw [dottedarrow] (linear) -- (terminalposition);
    \draw[dottedarrow](minloss) -- (S);
    \draw[dottedarrow] (suparg) -- (OCminarg);
\end{tikzpicture}
\end{adjustbox}

    \caption{(See Section~\ref{sec:theory}) Mathematical formulation describing the connection between a regularized learning problem with integral-type loss (\textbf{top}), the generalized Hopf formula for HJ PDEs with time-dependent Hamiltonians (\textbf{middle}), and the corresponding optimal control problem (\textbf{bottom}). The content of this illustration matches that of Figure~\ref{fig:intro_connection_in_words} by replacing each term in Figure~\ref{fig:intro_connection_in_words} with its corresponding mathematical expression. The colors indicate the associated quantities between each problem. The solid-line arrows denote direct equivalences. The dotted arrows represent additional mathematical relations.}
    \label{fig:connection_general}
\end{figure}

%\update{TO DO: change $F(s;\weightvec)$ to $\mathcal{F}_\weightvec(s)$ every time we type AF, get rid of "abuse of notation", just new notation}
In this section, we connect the Hopf formula~\eqref{eq:generalizedHopf} to regularized learning problems with integral-type losses (e.g. see \cite{sirignano2018dgm}). Consider a learning problem with data $(s,\by(s)) \in [0,t]\times \LPoutspace$. The goal is to learn a function $F(\cdot;\weightvec)$ with inputs in $[0,t]$ and unknown parameter $\weightvec\in\weightspace$. We learn $F(\cdot;\weightvec)$ as follows. First, we apply an operator $\mathcal{A}$ on the function $F(\cdot;\weightvec)$, 
where we denote $\mathcal{A}\mathcal{F}_\weightvec(s):=\mathcal{A}[F(\cdot;\weightvec)](s)$. %abuse notation and denote $\mathcal{A}F(s;\weightvec):=\mathcal{A}[F(\cdot;\weightvec)](s)$.  
For example, $\mathcal{A}$ could be the identity operator (as in regression problems; e.g., see \cite{weisberg2005applied}) or a differential operator (as in PINNs; e.g., see \cite{raissi2019physics}). 
The discrepancy between the learned model $\mathcal{A}\mathcal{F}_\weightvec$ and the measurements $\by$ is measured using an integral-type data fitting loss $\int_0^t\lossfunc(\mathcal{A}\mathcal{F}_\weightvec(s), \by(s))ds$, where we assume that the function $\weightvec \mapsto \lossfunc(\mathcal{A}\mathcal{F}_\weightvec(s), \by(s))$ is convex. For example, setting $\lossfunc(\ba, \bb) = \sum_{i=1}^m |a_i-b_i|$ or $\lossfunc(\ba, \bb) = \sum_{i=1}^m (a_i-b_i)^2$ yields an $L_1$ or $L_2$-squared data fitting loss, respectively. The unknown parameters $\weightvec$ are then learned by solving the following optimization problem:
\begin{equation}\label{eqt:general_learning}
\min_{\weightvec\in \weightspace} \lambda\int_0^t \lossfunc(\mathcal{A}\mathcal{F}_\weightvec(s), \by(s))ds + \regfunc(\weightvec),
\end{equation}
where $\lambda>0$ is a weight on the data fitting loss and $R$ is a convex regularization term.

Then, the connection between the learning problem~\eqref{eqt:general_learning}, the generalized Hopf formula~\eqref{eq:generalizedHopf}, and the optimal control problem~\eqref{eq:opt_ctrl} is summarized in Figure~\ref{fig:connection_general}. Specifically, the learning problem~\eqref{eqt:general_learning} is related to the generalized Hopf formula~\eqref{eq:generalizedHopf} by setting $\weightvec = \HJmom$, $\Hamiltonian(s,\HJmom) = \lambda\lossfunc(\mathcal{A}\mathcal{F}_\HJmom(s), \by(s))$, and $\regfunc(\HJmom) = \HJIC^*(\HJmom) - \langle \HJx,\HJmom\rangle + c(\HJx)$, where $c(\HJx)$ is a constant (possibly 0) that is independent of $\HJmom$ but may depend on $\HJx$. In other words, the variable $\HJx$ in the HJ PDE becomes a hyper-parameter in the learning problem, and we can treat it as a constant when optimizing the learning problem~\eqref{eqt:general_learning} with respect to $\weightvec = \HJmom$. Hence, when we solve these learning problems, we also solve the optimal control problem~\eqref{eq:opt_ctrl} with initial position $\HJx$ and time horizon $\HJt$, or, equivalently, we evaluate the solution to the corresponding HJ PDE~\eqref{eq:HJPDE_timedependent} at the point $(\HJx,\HJt)$. Conversely, when we solve the HJ PDE~\eqref{eq:HJPDE_timedependent}, the spatial gradient $\nabla_\HJx S(\HJx, \HJt)$ of the solution gives the minimizer $\weightvec^*$ of the learning problem~\eqref{eqt:general_learning}.

\section{Regularized linear regression problems with quadratic integral-type losses}\label{sec:method}
\def\regmat{A} 
\def\regcenter{\bb}

As a first exploration of the theoretical connection we developed in Section~\ref{sec:general_connection}, we consider the specific case of quadratic-regularized linear regression problems with quadratic integral-type losses. We begin by establishing the connection for this specific case. %. Namely, this learning problem is connected to an HJ PDE with quadratic time-dependent Hamiltonian and quadratic initial condition, which is also connected to an LQR problem. We 
We then leverage this connection to develop a new Riccati-based approach for solving this learning problem, which has potential computational and memory advantages in continual learning and big data settings.

\subsection{Problem formulation and connection to LQR problems}\label{sec:lin_reg_theory}

The learning problem is formulated as follows. Given data $(s, \by(s))\in [0,t]\times \LPoutspace$, the goal is to learn a linear prediction model $\Phi(\cdot)\weightvec$ such that $\mathcal{A}\Phi(s)\weightvec \approx \by(s), \forall s\in[0,t]$, where $\Phi(\cdot) = [\phi_1(\cdot), ..., \phi_n(\cdot)]\in\R^{m\times n}$ is the matrix whose columns are the basis functions $\phi_j:[0,t]\to\R^m$, $j=1, \dots, n$, $\weightvec = [\weight_1, \dots, \weight_n]^T\in\Rn$ are unknown trainable coefficients, and $\mathcal{A}$ is a linear operator (e.g., identity operator, differential operators), where we abuse notation and denote $\mathcal{A}\Phi(s) := [\mathcal{A}\phi_1(s), ..., \mathcal{A}\phi_n(s)]$. We learn $\weightvec$ by minimizing the following loss function:
\begin{equation}\label{eq:loss_function}
    \mathcal{L}(\weightvec) = \int_0^t\frac{\lambda}{2}\|\mathcal{A}\Phi(s)\weightvec - \MLy(s)\|_2^2 ds + \frac{1}{2}\|\regmat\weightvec - \regcenter\|_2^2,
\end{equation}
where $\frac{1}{2}\|\regmat\weightvec - \regcenter\|_2^2$ is a quadratic regularization term, $\regmat\in\R^{n\times n}$ is a positive definite matrix that weights the regularization, and $\bb\in\Rn$ acts as a prior on $\weightvec$ that biases $\weightvec$ to be close to $\regmat^{-1}\regcenter$. Note that this loss function~\eqref{eq:loss_function} is strictly convex; hence, it has a unique global minimizer. 

Then, this learning problem has connections with a generalized Hopf formula and an optimal control problem.
By expanding the square, we see that computing the minimizer of the loss function~\eqref{eq:loss_function} is equivalent to computing the maximizer of the following generalized Hopf formula:
\begin{equation}\label{eq:genHopf_quad}
\begin{aligned}
    S(\HJx, \HJt) & = \sup_{\HJmom\in\Rn}\left\{\langle\HJx,\HJmom\rangle- \int_0^t\frac{\lambda}{2}\|\mathcal{A}\Phi(s)\HJmom - \MLy(s)\|_2^2 ds - \frac{1}{2}\|\regmat\HJmom\|_2^2 + \langle\regmat\HJmom, \regcenter\rangle\right\},
\end{aligned}
\end{equation}
when $\HJx = 0$. In other words, solving this learning problem~\eqref{eq:loss_function} is equivalent to solving the HJ PDE~\eqref{eq:HJPDE_timedependent} with time-dependent, quadratic Hamiltonian $\Hamiltonian(s,\HJmom) = \frac{\lambda}{2}\|\mathcal{A}\Phi(s)\HJmom - \by(s)\|_2^2$ and quadratic initial condition $\HJIC(\HJx) = \frac{1}{2}\|(\regmat^{-1})^T\HJx + \regcenter\|_2^2$ at the point $(0,t)$. By extension, solving both of these problems is also equivalent to solving the optimal control problem~\eqref{eq:opt_ctrl} with running cost $L(s,\HJu) = \frac{1}{2}\HJu^T\HJu - \sqrt{\lambda}\by(s)^T\HJu$, terminal cost $\HJIC(\HJx) = \frac{1}{2}\|(\regmat^{-1})^T\HJx + \regcenter\|_2^2$, dynamics $f(s,\HJu) = \sqrt{\lambda}(\mathcal{A}\Phi(s))^T\HJu$, initial position $\bx(0) = 0$, and time horizon $\HJt$. Note that since this optimal control problem has quadratic running and terminal costs and linear dynamics, it is actually an LQR problem with time-dependent costs and dynamics (e.g., see \cite{anderson2007optimal, tedrake2023underactuated}).

\subsection{Riccati-based methodology}\label{sec:riccati}
It is well-known that LQR problems can be solved using Riccati ODEs (e.g., see~\cite{mceneaney2006max}). By our connection, this gives us that the learning problem~\eqref{eq:loss_function} can also be solved using Riccati ODEs. 
Namely, the viscosity solution $S$ (given by~\eqref{eq:genHopf_quad}) to the corresponding HJ PDE is also given by $S(\HJx,t) = \frac{1}{2}\HJx^T\Sxx(t)\HJx + \Sx(t)^T\HJx + \Sc(t)$, where 
$\Sxx:[0,\infty)\to\R^{n\times n}$, which takes values in the space of positive definite matrices, $\Sx:[0,\infty)\to\Rn$, and $\Sc:[0,\infty)\to\R$ 
satisfy the following Riccati ODEs:
\begin{equation}\label{eqt:RiccatiODEs}
    \begin{dcases}
    \dot{\Sxx}(s) =  -\lambda\Sxx(s)^T(\mathcal{A}\MLbasismat(s))^T\mathcal{A}\MLbasismat(s)\Sxx(s) &s>0,\\
    \dot{\Sx}(s) = -\lambda\Sxx(s)^T(\mathcal{A}\MLbasismat(s))^T(\mathcal{A}\MLbasismat(s)\Sx(s) - \MLy(s))&s>0, \\
    \dot{\Sc}(s) = -\frac{\lambda}{2}\left\|\mathcal{A}\MLbasismat(s)\Sx(s)- \by(s)\right\|_2^2 & s > 0
    \end{dcases}
\end{equation}
with initial conditions $\Sxx(0) = \regmat^{-1}(\regmat^{-1})^T$, $\Sx(0) = \regmat^{-1}\regcenter$, and $\Sc(0) = \frac{1}{2}\regcenter^T\regcenter$. Then, the minimizer $\weightvec^*$ of the learning problem~\eqref{eq:loss_function} is given by 
%\begin{equation}%\label{eq:optimal_coeff}
 $   \weightvec^* = \HJmom^* = \nabla_\HJx S(0,t) = \Sx(t),$ 
%\end{equation}
where $\HJmom^*$ is the optimizer in the generalized Hopf formula~\eqref{eq:genHopf_quad} when $\HJx = 0$. 

\begin{rem}
When solving the learning problem~\eqref{eq:loss_function}, we only care about computing the minimizer $\weightvec^*$ and not the minimal value of the loss function. Hence, solving the learning problem using the Riccati ODEs~\eqref{eqt:RiccatiODEs} only requires solving the ODEs for $\Sxx,\Sx$, and the ODE for $\Sc$ can be ignored. In the remainder of the paper, we disregard the ODE for $\Sc$.
\end{rem}

Solving the learning problem~\eqref{eq:loss_function} via the Riccati ODEs~\eqref{eqt:RiccatiODEs} provides computational and memory advantages in certain learning scenarios, particularly those involving continual learning and/or very large datasets. For example, consider the case where our model has already been trained using the data $(s, \by(s))\in [0,t]\times \LPoutspace$ and we want to incorporate some additional data points $(s, \by(s))\in (t,t_1]\times \LPoutspace$, where $t_1 > t$, into our learned model. Then, our model can be updated by simply evolving the Riccati ODEs~\eqref{eqt:RiccatiODEs} from $t$ to $t_1$. Note that doing so requires neither retraining on the entire dataset $(s, \by(s))\in [0,t_1]\times \LPoutspace$ nor access to any of the previous data $(s, \by(s))\in [0,t]\times \LPoutspace$ since evolving the Riccati ODEs~\eqref{eqt:RiccatiODEs} from $t$ to $t_1$
only requires the values $\Sxx(t),\Sx(t)$ (i.e., the results of the previous training) to initialize the Riccati ODEs and the new data $(s, \by(s))\in (t,t_1]\times \LPoutspace$ to continue evolving the ODEs on the subsequent time interval $[t,t_1]$. Hence, this Riccati-based approach is particularly beneficial in continual learning scenarios, wherein learned models are incrementally updated as new data becomes available (and hence, constantly retraining on the entire, growing dataset can quickly become computationally infeasible), and memory-constrained learning scenarios where datasets are too large to store in their entirety.  

Now consider the case where we want to remove some data $(s, \by(s))\in [t_2,t]\times \LPoutspace$, where $t_2 < t$, which has already been incorporated into our learned model. This scenario may correspond to the case where we want to remove some corrupted data in order to improve the accuracy of the learned model. Then, this data can be removed by reversing time and solving the Riccati ODEs~\eqref{eqt:RiccatiODEs} from $t$ to $t_2$ with terminal conditions given by $\Sxx(t), \Sx(t)$. As in the previous case, the advantages of this Riccati-based approach are that we only require access to the data that is being removed and not the entire dataset. Thus, if we are removing a small amount of data (relative to the size of the entire dataset), this approach can be more efficient than retraining on the entire remaining dataset.

\begin{rem}
We have flexibility in how we connect the learning problem~\eqref{eq:loss_function} to an HJ PDE. Specifically, in~\eqref{eq:genHopf_quad}, we interpret $\regcenter$ as part of the initial condition $J$ of the corresponding HJ PDE. However, we could also interpret $\regcenter$ to be part of $\HJx$, the point at which we evaluate the HJ PDE. Let $\regcenter = \regcenter_\HJIC + \regcenter_\HJx$, where $\regcenter_\HJIC$ and $\regcenter_\HJx$ are the parts of $\regcenter$ that will be associated with $\HJIC$ and $\HJx$, respectively. Then, the new corresponding HJ PDE and optimal control problem are identical to those defined in Section~\ref{sec:lin_reg_theory} but with $\regcenter$ replaced with $\regcenter_\HJIC$ and evaluated at $\HJx =\bx(0)=\regmat^T\regcenter_\HJx$ (instead of at $\HJx =\bx(0)=0$). Let $\tilde S$ be the solution to this new HJ PDE. Then, the minimizer $\weightvec^*$ of the learning problem~\eqref{eq:loss_function} can alternatively be computed as
%\begin{equation}\label{eq:optimal_weights_alternative}
 $   \weightvec^* = \HJmom^* = \nabla_\HJx \tilde S(\regmat^T\regcenter_\HJx,t) = \tilde\Sxx(t)\regmat^T\regcenter_\HJx + \tilde\Sx(t),$ 
%\end{equation}
where $\tilde \Sxx, \tilde\Sx$ satisfy the Riccati ODEs~\eqref{eqt:RiccatiODEs} with initial conditions $\Sxx(0) = \regmat^{-1}(\regmat^{-1})^T$, $\Sx(0) = \regmat^{-1}\regcenter_\HJIC$. The advantage of this alternative interpretation is that $\regcenter_\HJx$ (i.e., the bias on $\weightvec$) can be tuned efficiently using only 2 matrix-vector multiplications and 1 vector addition %according to the above equation for $\weightvec^*$ %~\eqref{eq:optimal_weights_alternative} 
instead of having to re-solve the Riccati ODEs~\eqref{eqt:RiccatiODEs} with a new initial condition (i.e., re-training on the entire dataset with a new bias). Note that when $\regcenter_\HJx = 0$, we recover the original framework from Section~\ref{sec:lin_reg_theory}.
\end{rem}

\section{Numerical examples}\label{sec:examples}

% {\color{blue}Two computational examples are presented in this section. We use least squares estimate (LSE) as the reference method, in which the integral-type loss is approximated by Monte Carlo method. We note that the reference method requires the information of $\mathcal{A}\Phi$ and $\MLy$ on the whole domain.}

In this section, we apply the Riccati-based methodology from Section~\ref{sec:method} to two test problems to demonstrate the potential computational and memory advantages of our approach. For illustration purposes, we apply 4th-order Runge-Kutta (RK4) to solve the Riccati ODEs in each example. Note that our methodology does not rely on any particular numerical solver. Hence RK4 could be replaced by any other appropriate numerical method. Code for all examples will be made publicly available at \url{https://github.com/ZongrenZou/TimeHJPDE4SciML} after this paper is accepted.

\subsection{A boundary-value ODE problem}

\begin{figure}[ht]
    \centering
    \subfigure[Inference of $u$ after ${[0, t]} \ni s\mapsto  f(s)$ is visited.]{
        \includegraphics[scale=.35]{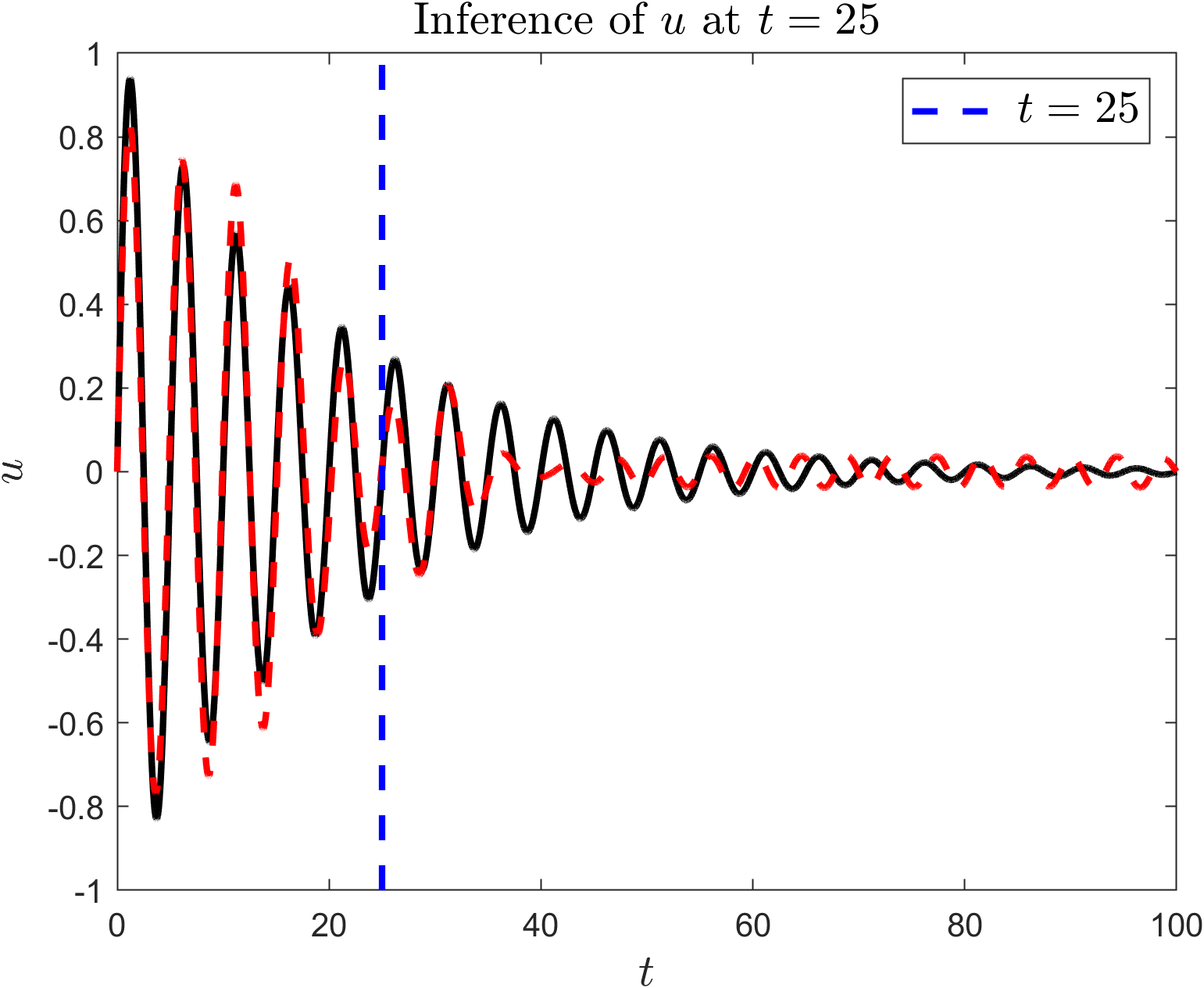}
        \includegraphics[scale=.35]{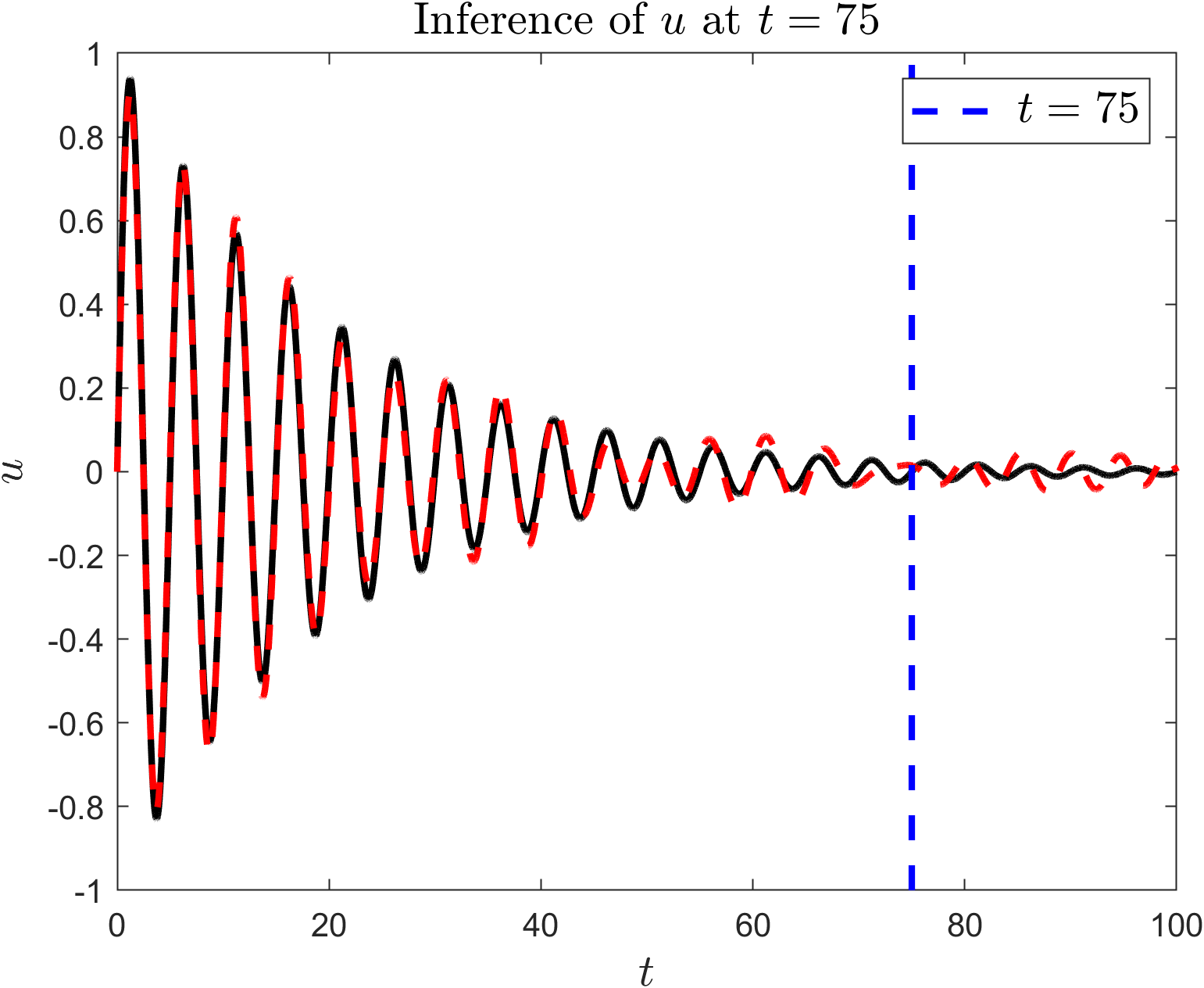}
        \includegraphics[scale=.35]{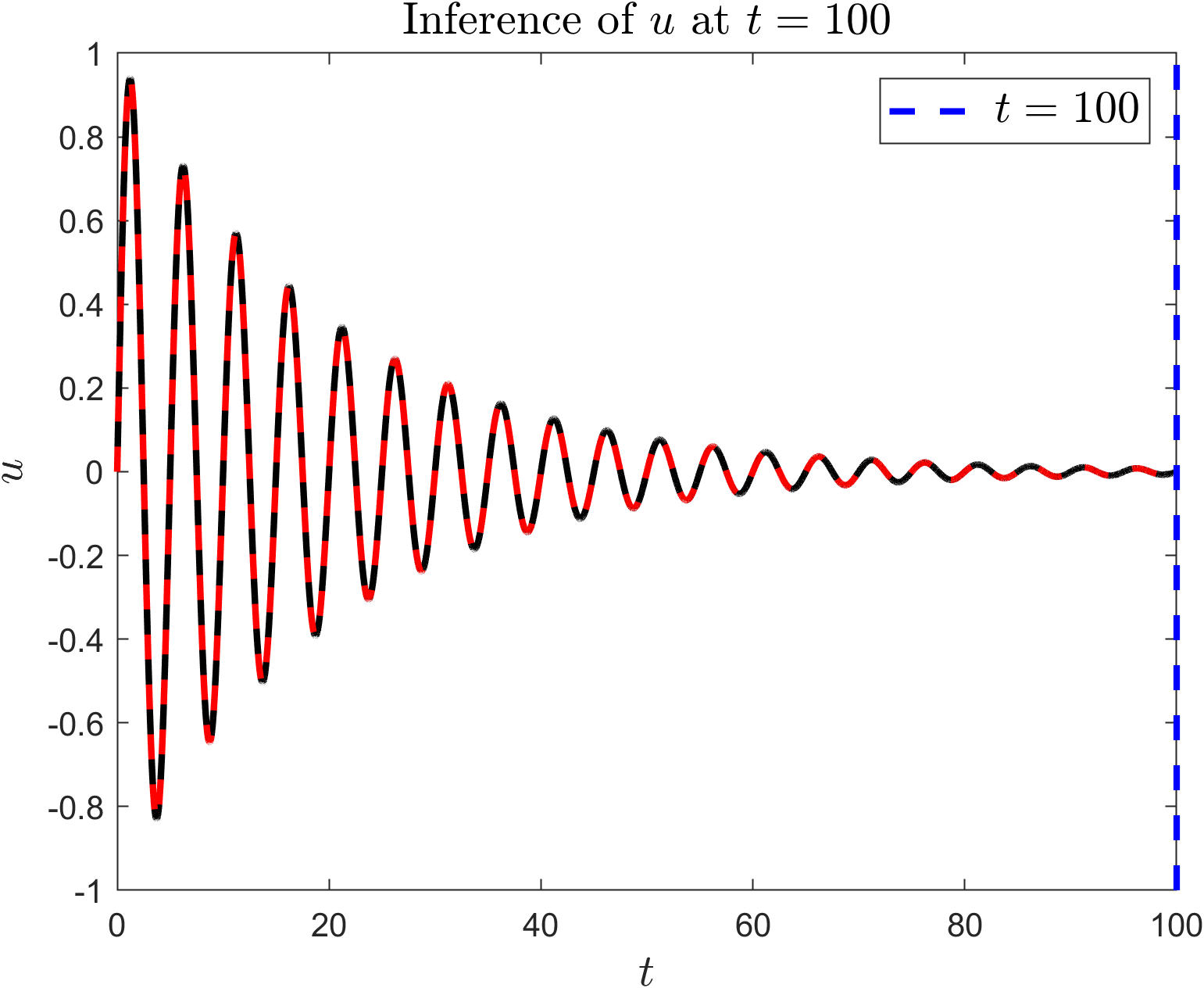}
    }
    \subfigure[Inference/fitting of $f$ after ${[0, t]} \ni s\mapsto f(s)$ is visited.]{
        \includegraphics[scale=.35]{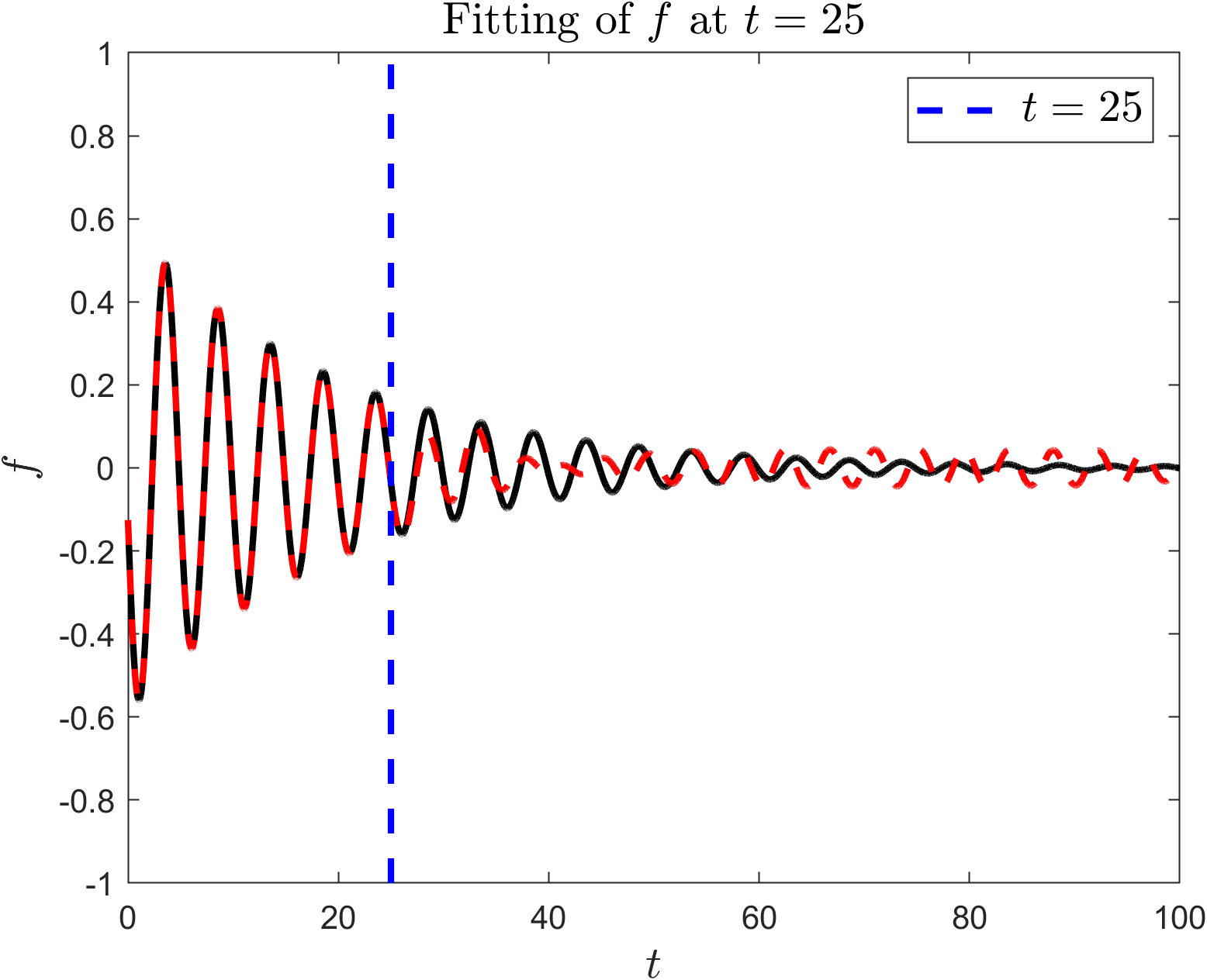}
        \includegraphics[scale=.35]{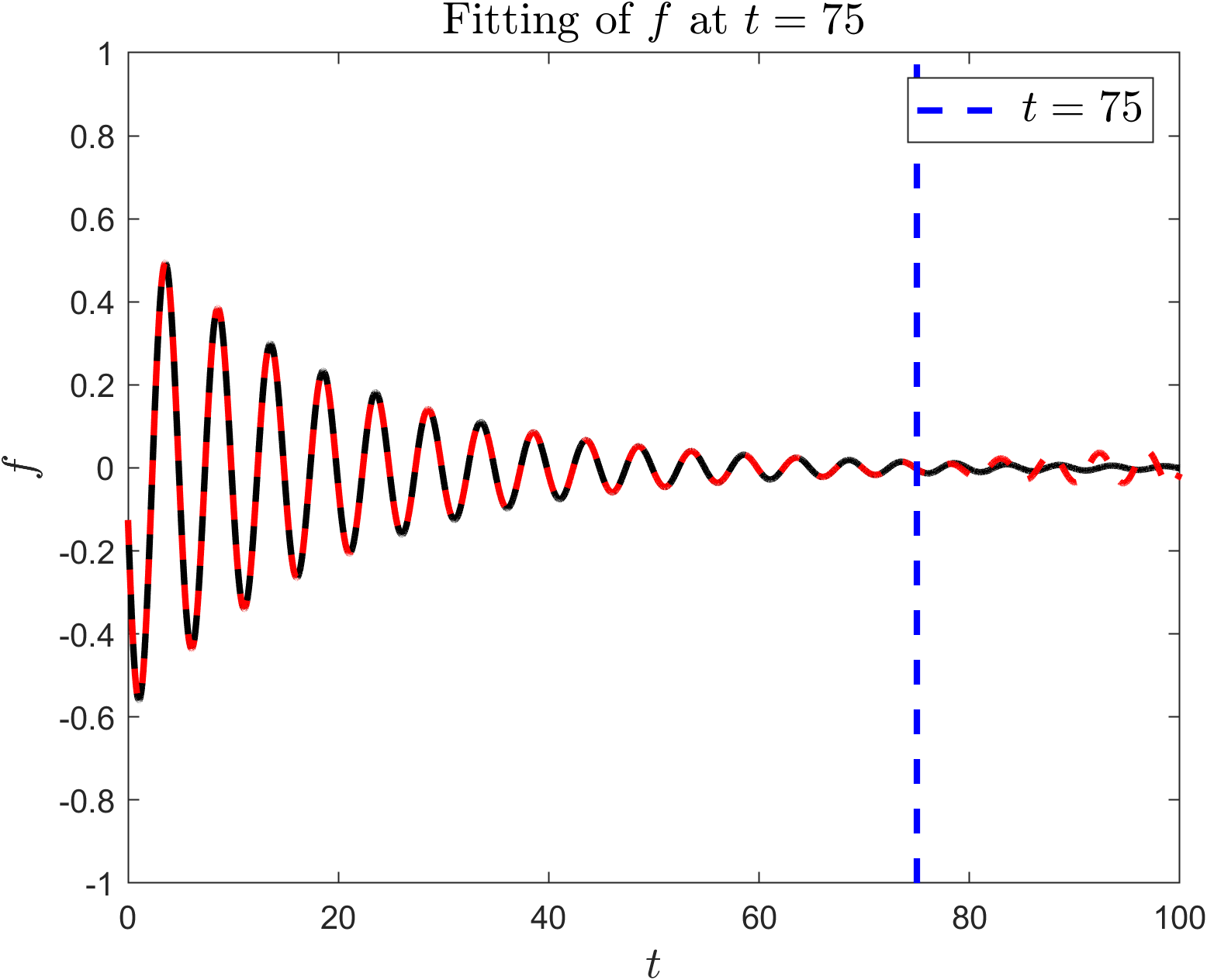}
        \includegraphics[scale=.35]{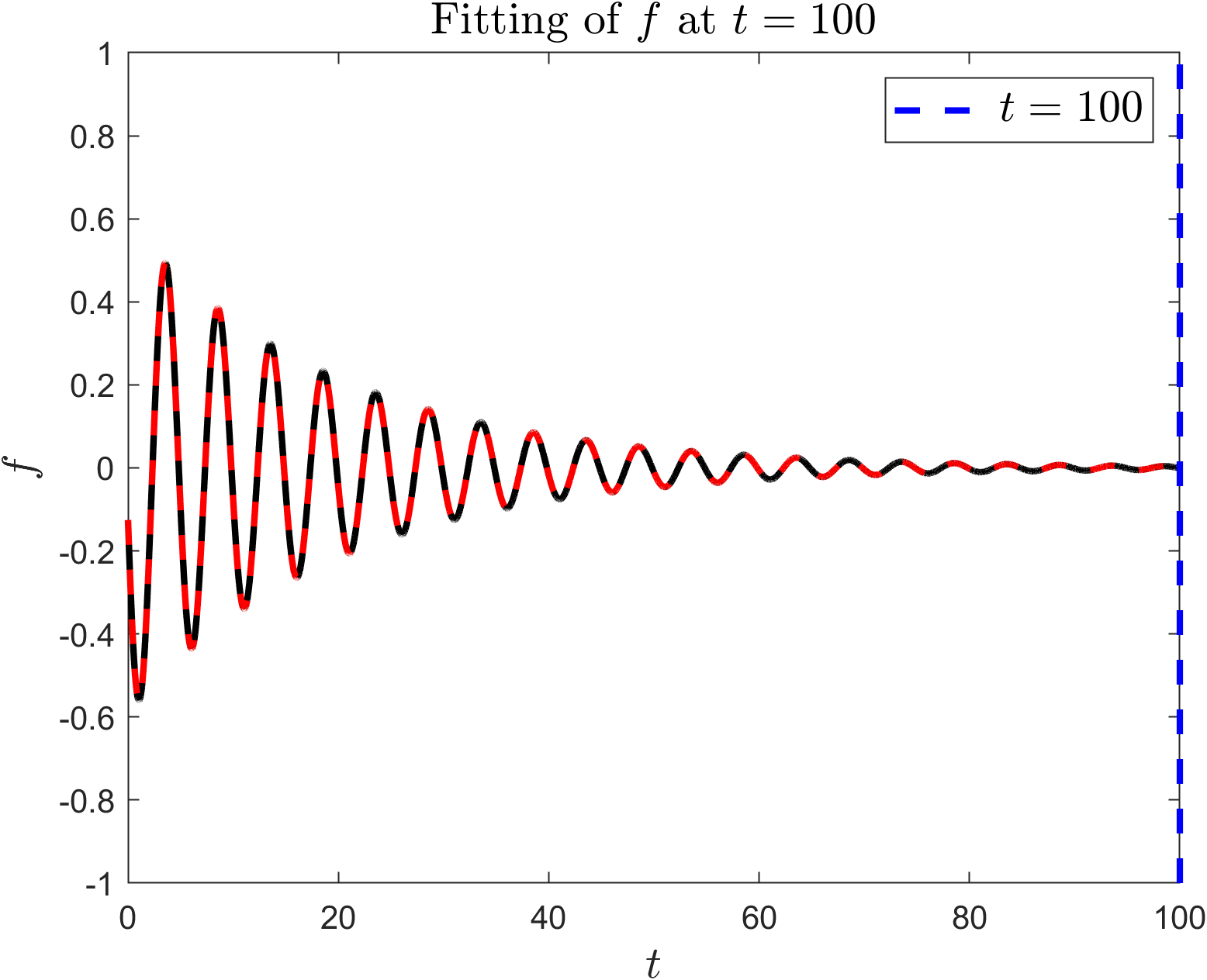}
    }
    \caption{Continual learning of the solution $u$ and source term $f$ of the boundary-value ODE \eqref{eq:ode} using our Riccati-based approach, where information of $f$ is treated as a flow with respect to $t$ and cannot be stored once visited. \textcolor{red}{\textbf{$--$}}: inferences of $u$, $f$ at different $t$; \textcolor{black}{\textbf{---}}: exact values of $u$, $f$; \textcolor{blue}{\textbf{$--$}}: where integration has advanced in $t$ so far. %Quantitative results can be found in Table \ref{tab:example1}. 
    Our Riccati-based approach naturally coincides with the continual learning framework by allowing new information to be continuously incorporated into the learned model without requiring access to any of the previous information. Instead, all of the previous information is encoded in the solution to the corresponding HJ PDE, thus avoiding catastrophic forgetting.}
    \label{fig:example1}
\end{figure}

\noindent In this example, we apply our Riccati-based methodology to solve a boundary value problem using continual learning (\cite{parisi2019continual, van2019three}) to demonstrate the potential computational and memory benefits of our approach. %Namely, this example demonstrates how our Riccati-based approach naturally matches the continual learning framework, while inherently avoiding catastrophic forgetting.
Consider the following ODE problem:
\begin{equation}\label{eq:ode}
    \begin{dcases}
        \frac{d^2u}{dt^2} + u(t) = f(t), t\in(0, T),
    \\ u(0) = u_0, u(T) = u_T,
    \end{dcases}
\end{equation}
where $f:[0,T]\to\R$ is the source term derived from $u(t) = \exp(-0.05t)\sin(0.4\pi t)$, $t\in[0, T]$ (although $u$ is assumed to be unknown a priori), $u_0=0$ and $u_T=\exp(-0.05T)\sin(0.4\pi T)$ are constants, and $T = 100$ so that this is a long-term integration problem. 
%and $u_0,u_T\in\R$ are constants.
%In this example, we consider a long-term integration problem where $T$ is relatively large.
Following the continual learning framework, we assume that information of $f$ is accessed in a stream as a flow of $t$ and that information of $f$ becomes inaccessible after it is incorporated into our learned model. % and therefore is only available in a stream, making it a continual learning problem \cite{parisi2019continual}.
Our goal is %to use this information of $f$ 
to learn the linear model $\sum_{i=1}^n\weight_i \MLbasis_i(\cdot)$ to approximate the solution $u$ of the ODE~\eqref{eq:ode}, where $\{\MLbasis_i(\cdot)\}_{i=1}^n = \{1\} \cup \{j\sin(j\cdot), j\cos(j\cdot)\}_{j=1}^{\frac{n-1}{2}}, n=301$. We learn the unknown coefficients $\weightvec = [\weight_1, \dots, \weight_n]^T$  using the following PINN-type loss (\cite{raissi2019physics, zou2023correcting}):
\begin{equation}\label{eq:loss_ode}
%\begin{adjustbox}{width=\textwidth}
\begin{split}
    \mathcal{L}(\weightvec; t) &= 
    \frac{\lambda_f}{2} \int_0^t \left|\sum_{i=1}^n\weight_i \left(\frac{d^2 \MLbasis_i }{d t^2}(s) + \MLbasis_i(s)\right) - f(s)\right|^2 ds \\ & \hspace{1in}
    + \frac{\lambda_0}{2} \left|\sum_{i=1}^n\MLbasis_i(0)\weight_i - u_0\right|^2 + \frac{\lambda_T}{2} \left|\sum_{i=1}^n\MLbasis_i(T)\weight_i - u_T\right|^2 + \frac{1}{2}\sum_{i=1}^n \gamma_i |\theta_i|^2,  
\end{split}
%\end{adjustbox}
\end{equation}
%We solve this problem using the Riccati-based approach brought by the connection between time-dependent HJ PDE and SciML. Specifically, we assume a linear surrogate model $u_\theta(\tau) = \sum_{k=1}^n\weight_k \MLbasis_k(\tau)$ ($\{\MLbasis_k(\tau)\}_{k=1}^n$ are basis functions and $\{\weight_k\}_{k=1}^n$ are their coefficients) and follow the PINN framework \cite{raissi2019physics, zou2023correcting} to establish the following loss function:
%\begin{equation}\label{eq:loss_ode}
%    \mathcal{L}(\weightvec; s) = \lambda_f\int_0^s |\frac{d^2u_\weightvec}{dt^2}(\tau) + u_\weightvec(\tau) - f(\tau)|^2 d\tau + \lambda_0 |u_\weightvec(0) - u_0|^2 + \lambda_T |u_\weightvec(T) - u(T)|^2 + \sum_{k=1}^n \gamma_k |\theta_k|^2,
%\end{equation}
where $\lambda_f>0$ is the belief weight for the ODE, $\lambda_0, \lambda_T\geq 0$ are belief weights for the boundary conditions, and $\gamma_i >0, i=1,\dots,n$ are belief weights for the $\ell^2$-regularization. 
Following the data streaming paradigm, to achieve real-time inferences, we need to minimize $\mathcal{L}(\cdot; t)$ for all $t\in[0, T]$. %Plugging $u_\theta(t) = \sum_{k=1}^n\weight_k \MLbasis_k(t)$ into \eqref{eq:loss_ode} gives us
%\begin{equation}
%\begin{split}
    %\mathcal{L}(\weightvec; s) &= \lambda_f \int_0^s |\sum_{k=1}^n\weight_k (\frac{d^2 \MLbasis_k }{d t^2}(\tau) + \MLbasis_k(\tau)) - f(\tau)|^2 d\tau + \\
    %&\lambda_0 |\sum_{k=1}^n\MLbasis_k(0)\weight_k - u_0|^2 + \lambda_T |\sum_{k=1}^n\MLbasis_k(T)\weight_k - u_T|^2 + \sum_{k=1}^n \gamma_k |\theta_k|^2.
%\end{split}
%\end{equation}
%We let $\Phi(s) = [\phi_1(s), \dots, \phi_n(s)]^T$, $\mathcal{A}\Phi(s) = [\frac{d^2\phi_1}{d\tau^2}(s) + \phi_1(s), \dots, \frac{d^2\phi_n}{d\tau^2}(s) + \phi_n(s)]^T$, $\MLy(s)=[f(s)]^T$, and $\Gamma$ be diagonal matrix whose $i$-th entry is $\gamma_i$. 
Note that by completing the square, minimizing the above loss function~\eqref{eq:loss_ode} is equivalent to minimizing a loss function in the form of~\eqref{eq:loss_function}, where $\lambda = \lambda_f$, $\Phi(\cdot) = [\phi_1(\cdot), \dots, \phi_n(\cdot)]$, $\mathcal{A}\Phi(\cdot) = [\frac{d^2\phi_1}{dt^2}(\cdot) + \phi_1(\cdot), \dots, \frac{d^2\phi_n}{dt^2}(\cdot) + \phi_n(\cdot)]$, $\MLy=f$, 
%\update{todo: check.. think i need to reverse where the transpose is} 
$\regmat = (\lambda_0\Phi(0)^T\Phi(0) + \lambda_T\Phi(T)^T\Phi(T) + \Gamma)^{1/2}$, where $\Gamma$ is the diagonal matrix whose $i$-th entry is $\gamma_i$, and $\regcenter = A^{-1}(\lambda_0\Phi(0)^Tu_0 + \lambda_T\Phi(T)^Tu_T)$. In other words, we treat the sum of the boundary and regularization terms in~\eqref{eq:loss_ode} as the regularization term in~\eqref{eq:loss_function}. Thus, the learning problem~\eqref{eq:loss_ode} can be solved using the Riccati ODEs~\eqref{eqt:RiccatiODEs}.  %Thus, this learning problem~\eqref{eq:loss_ode} can be solved using the Riccati ODEs~\eqref{eqt:RiccatiODEs} with initial conditions $\Sxx(0) = (\lambda_0\Phi(0)\Phi(0)^T + \lambda_T\Phi(T)\Phi(T)^T + \Gamma)^{-1}$, $\Sx(0) = (\lambda_0\Phi(0)\Phi(0)^T + \lambda_T\Phi(T)\Phi(T)^T + \Gamma)^{-1}(\lambda_0\Phi(0)u_0 + \lambda_T\Phi(T)u_T)$. 

For our numerics, we use $\lambda_f=100, \lambda_0 = \lambda_T = \gamma_i=1, i=1,...,n$, and RK4 with step size $h=10^{-6}$.
Note that evolving the Riccati ODEs from $t$ to $t+h$ only requires information of $f(s), s\in[t,t+h]$. Hence, our approach naturally coincides with the continual learning framework. 
%Meanwhile, all of the information from $\tau<t$ is inherently encoded in the solution to the corresponding HJ PDE via the solutions $\Sxx(t), \Sx(t)$ to the Riccati ODEs, which allows our approach to avoid catastrophic forgetting. 
% In Figure \ref{fig:example1} and Table \ref{tab:example1}, we observe that our inferences of both $u$ and $f$ improve as more information is incorporated into our learned model. In particular, we see that our inferences improve in accuracy in both the regions we have already visited and at future times, which indicates that catastrophic forgetting has not occurred.
%{\color{red}
In Figure \ref{fig:example1}, we observe that our inferences of both $u$ and $f$ improve as more information is incorporated into our learned model. 
In particular, we see that our inferences improve in accuracy in both the regions we have already visited and at future times, which indicates that catastrophic forgetting has not occurred. This behavior is consistent with the fact that all of the information from $s<t$ is inherently encoded in the solution to the corresponding HJ PDE via the solutions $\Sxx(t), \Sx(t)$ to the Riccati ODEs. 
Specifically, the relative $L_2$ errors of the inferences of $u$ are $27.13\%$, $12.36\%$, $0.03\%$ once $t = 25, 75, 100$ has been visited, respectively. Similarly, the relative $L_2$ errors of the inferences of $f$ are $23.83\%$, $8.08\%$, $0.03\%$ once $t = 25, 75, 100$ has been visited, respectively. 
For comparison, we also compute the least squares estimate (LSE) for \eqref{eq:loss_ode}, in which the integral is approximated using Monte Carlo with $10^6$ uniform sampling points. The relative $L_2$ errors of the LSE inferences of $u$ and $f$ are both $0.03\%$. We compute all of the above $L_2$ errors using trapezoidal rule with a uniform grid of size 1001 over the whole domain $[0,100]$. Note that LSE does not meet the requirements of continual learning as it requires access to the full information of $f$ at once (i.e., LSE requires information of $f(t), \forall t\in[0,100]$), yet our approach yields similar error levels once $t = 100$ has been visited even though we only access a small portion of the data at a time.
%The relative $L_2$ errors of the inferences of $u$ and $f$ are decreasing from $27.13\%$ and $23.83\%$ at $\tau=25$, to $12.36\%$ and $8.08\%$ at $\tau=75$, and $0.03\%$ and $0.03\%$ at $\tau=100$. 
%As a comparison, the least squares estimate (LSE) is employed, in which the integral-type loss is approximated using the Monte Carlo method with $10^6$ uniform sampling points, and the errors are $0.03\%$ and $0.03\%$. We note that LSE does not meet the requirements of continual learning as it requires access to the full information of $f$ at once.}
We also note that the inferences made using our approach can be considered in real time as they are updated every $h$ time units of data. Thus, our approach provides potential computational and memory advantages as updating the learned model using information on $[t,t+h]$ does not require retraining on or storage of any of the previous information from $[0,t)$.

\subsection{2D Poisson equation}\label{sec:examples_pde}

\begin{figure}[ht]
    \centering
    \subfigure[Absolute errors of the inference of $u$ after ${[0,1]\times [0, y]}\ni {(x,t)} \mapsto f(x,t)$ is visited.]{
        \includegraphics[scale=.3]{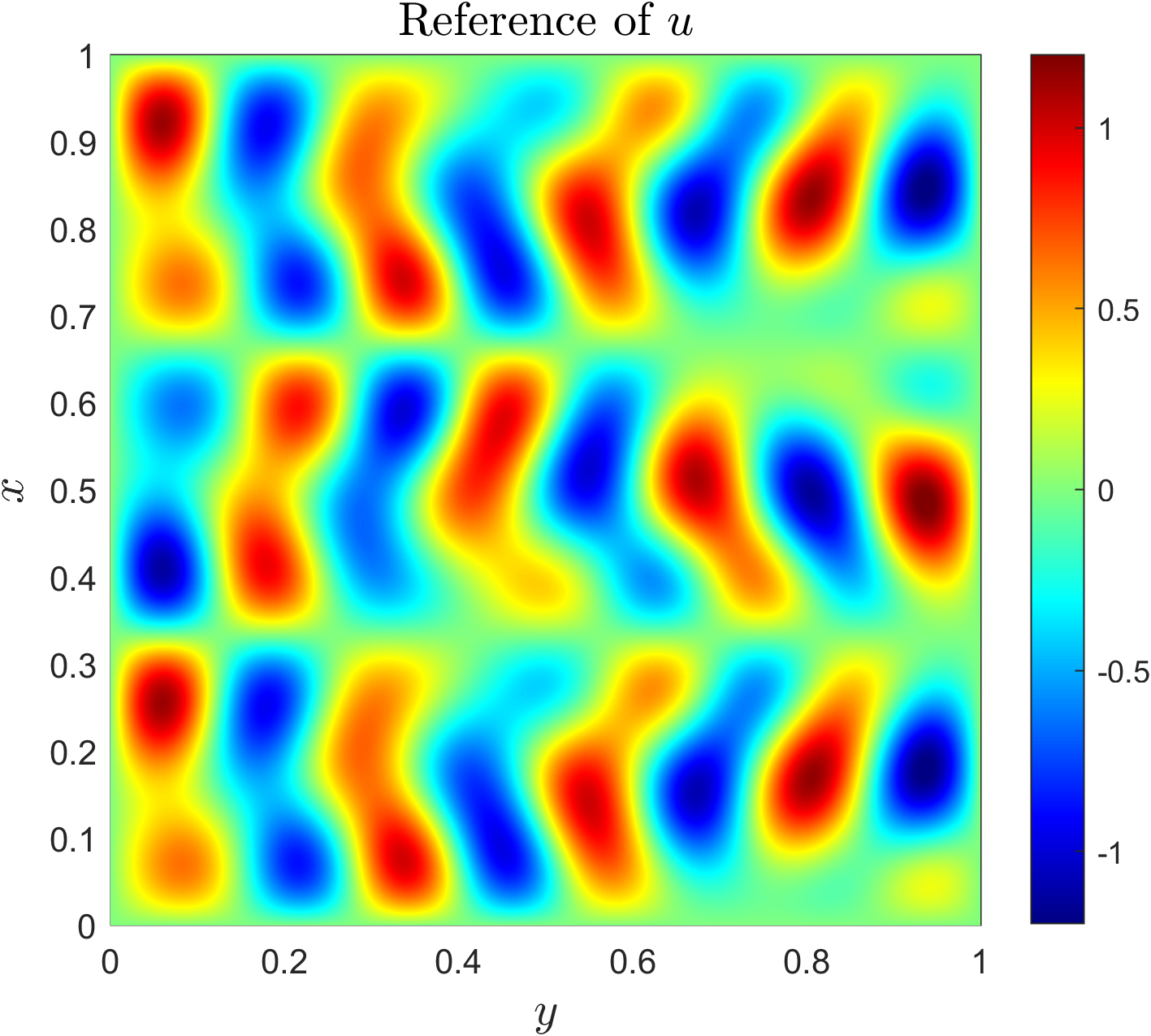}
        \includegraphics[scale=.3]{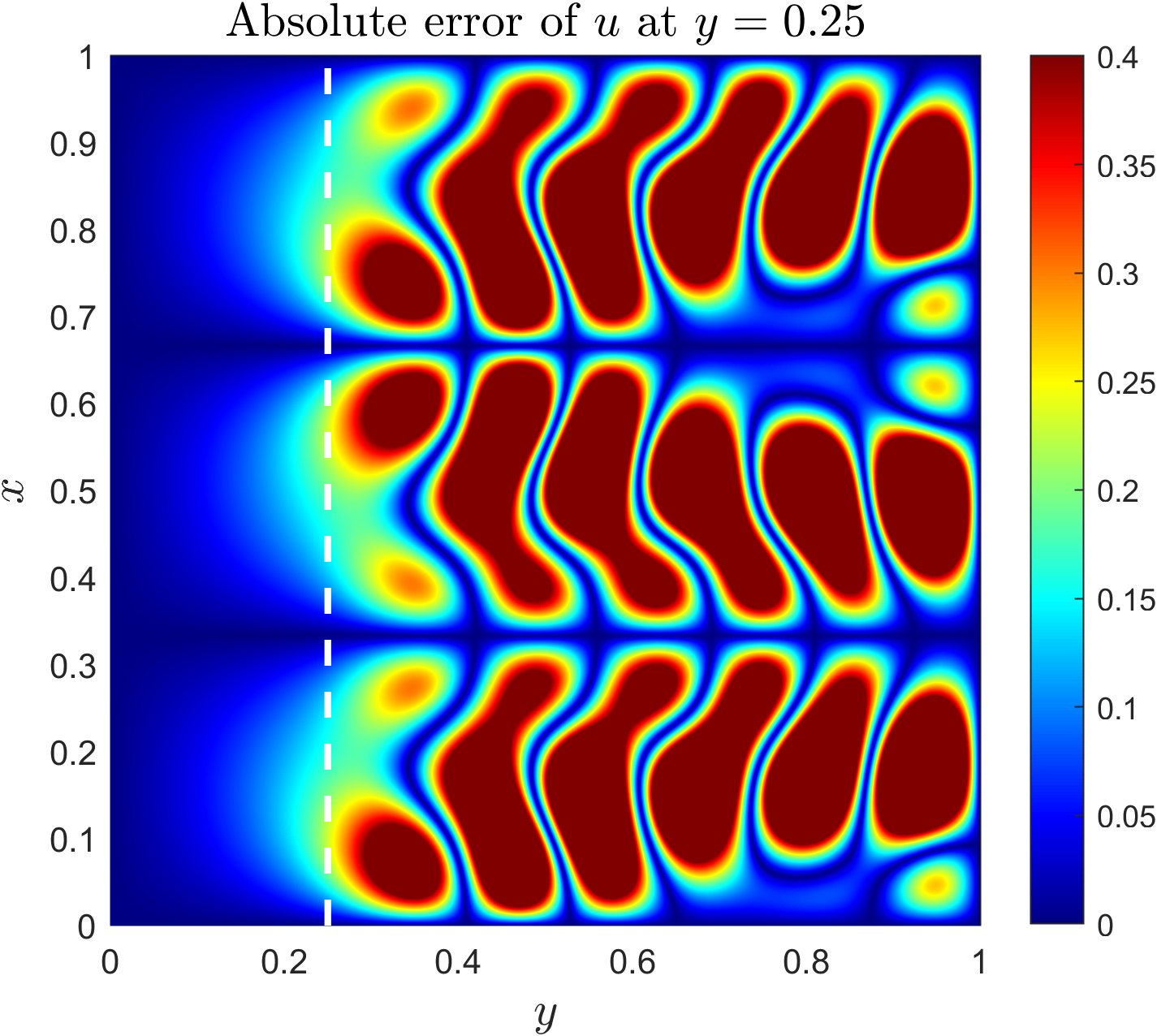}
        \includegraphics[scale=.3]{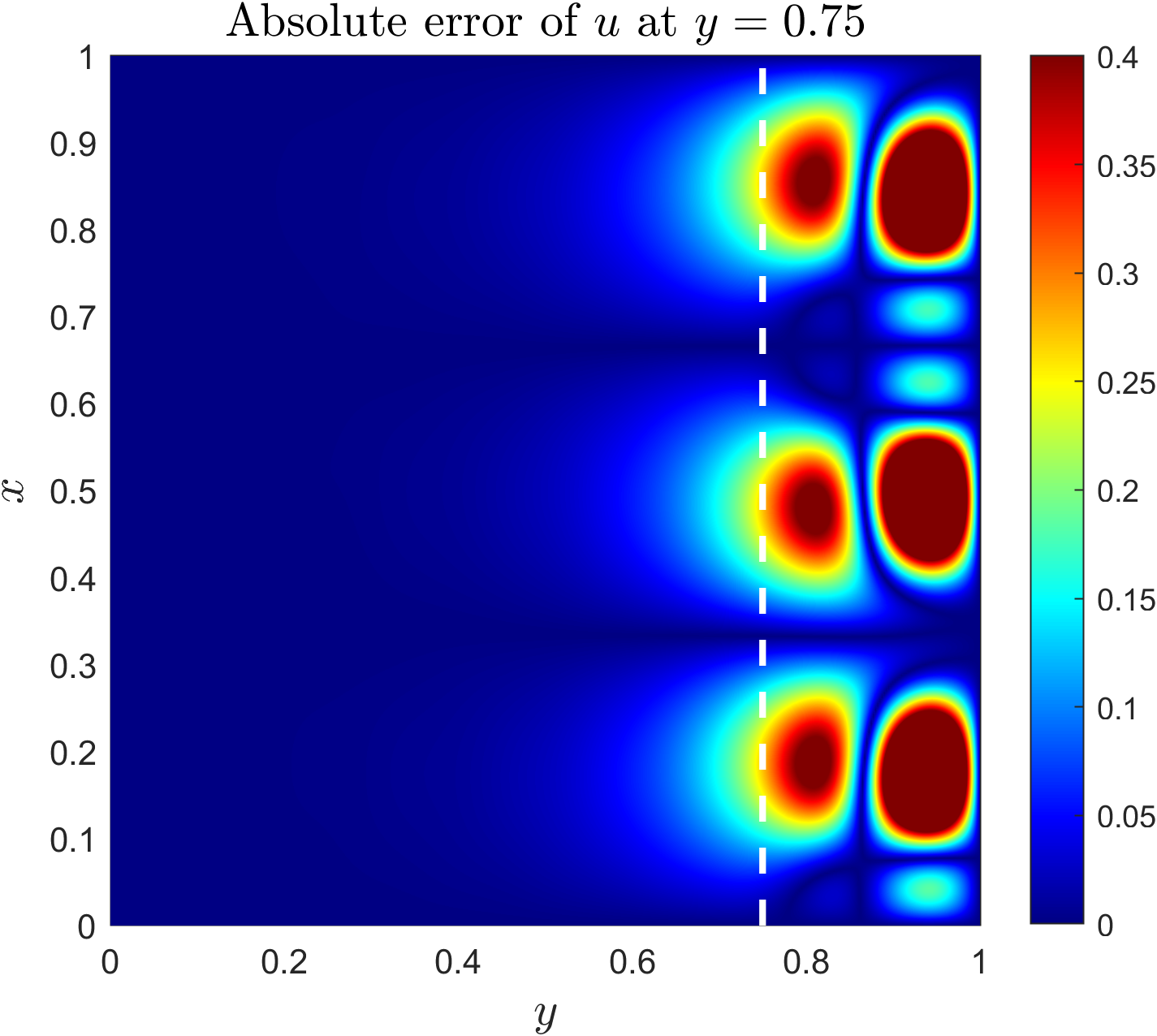}
        \includegraphics[scale=.3]{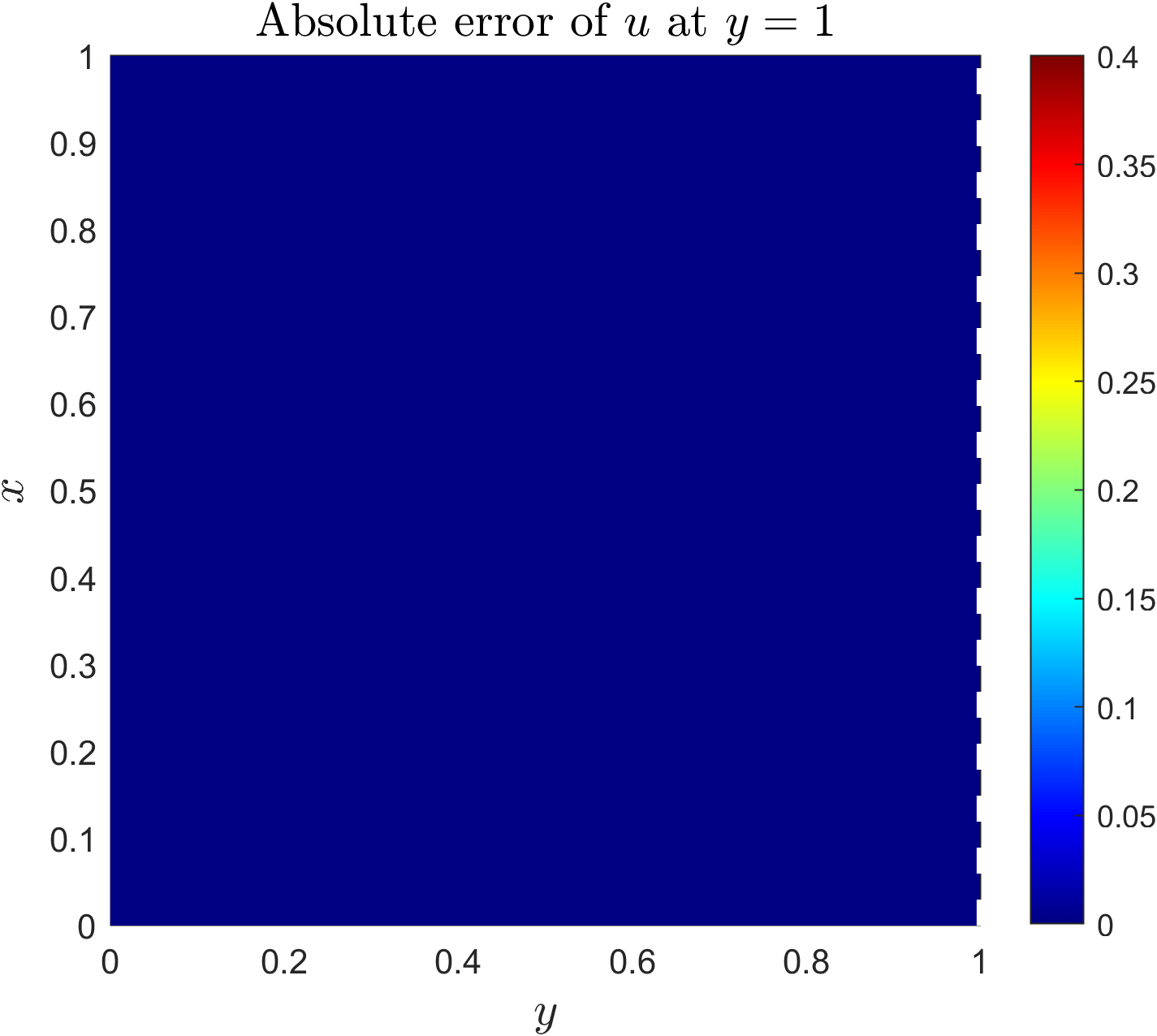}
    }
    \subfigure[Absolute errors of the inference/fitting of $f$ after ${[0,1]\times [0, y]}\ni {(x,t)} \mapsto f(x,t)$ is visited.]{
        \includegraphics[scale=.3]{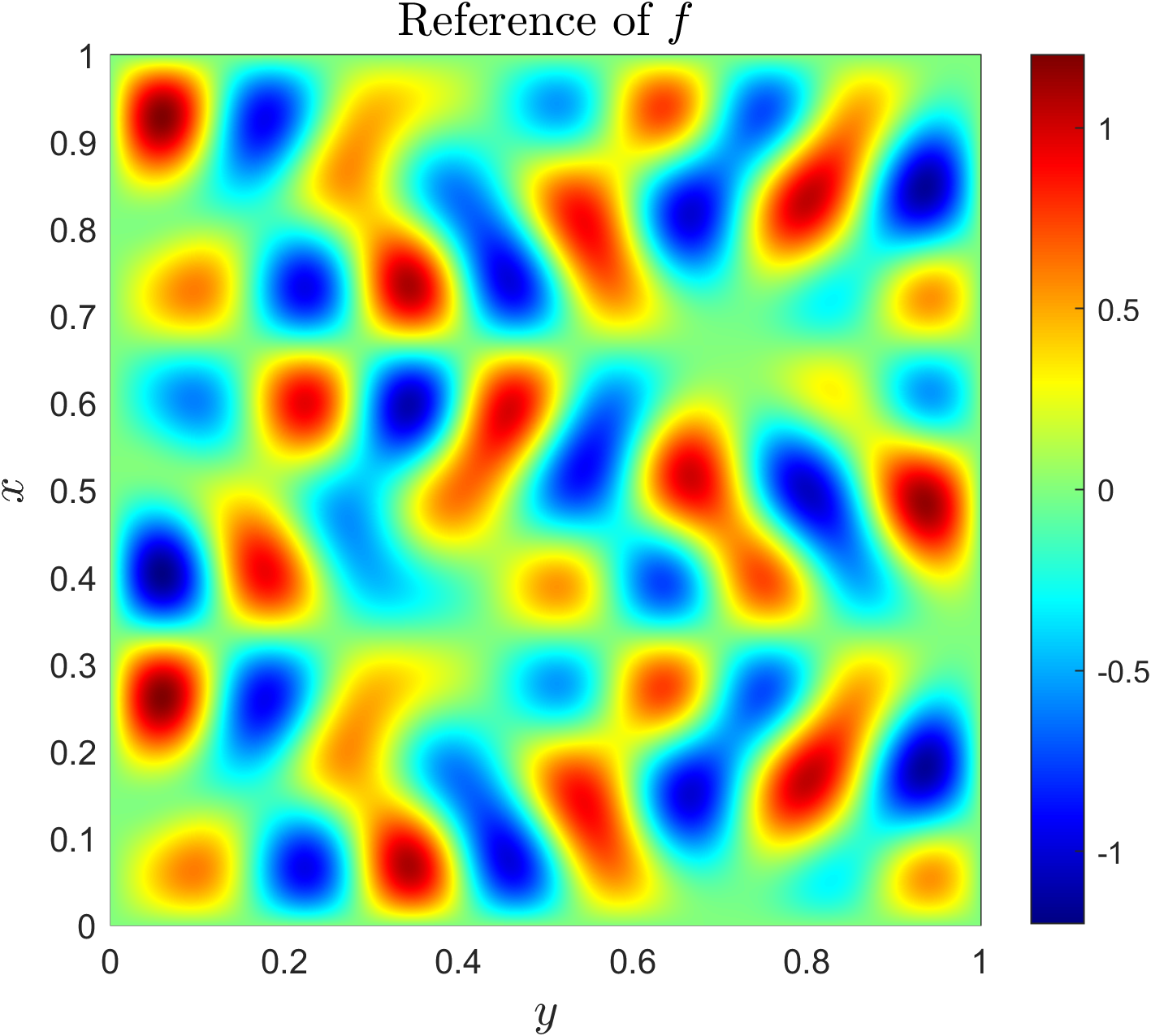}
        \includegraphics[scale=.3]{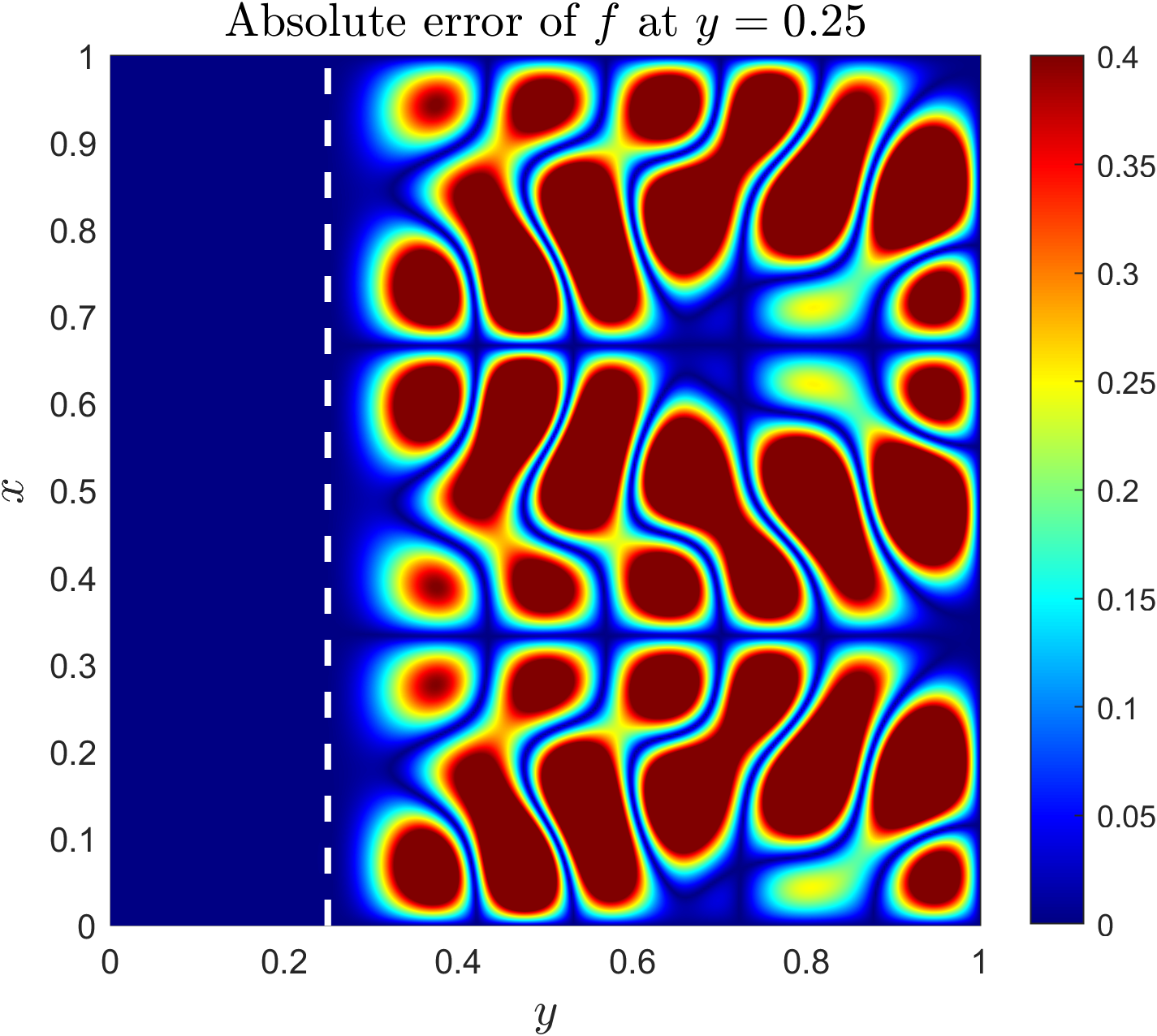}
        \includegraphics[scale=.3]{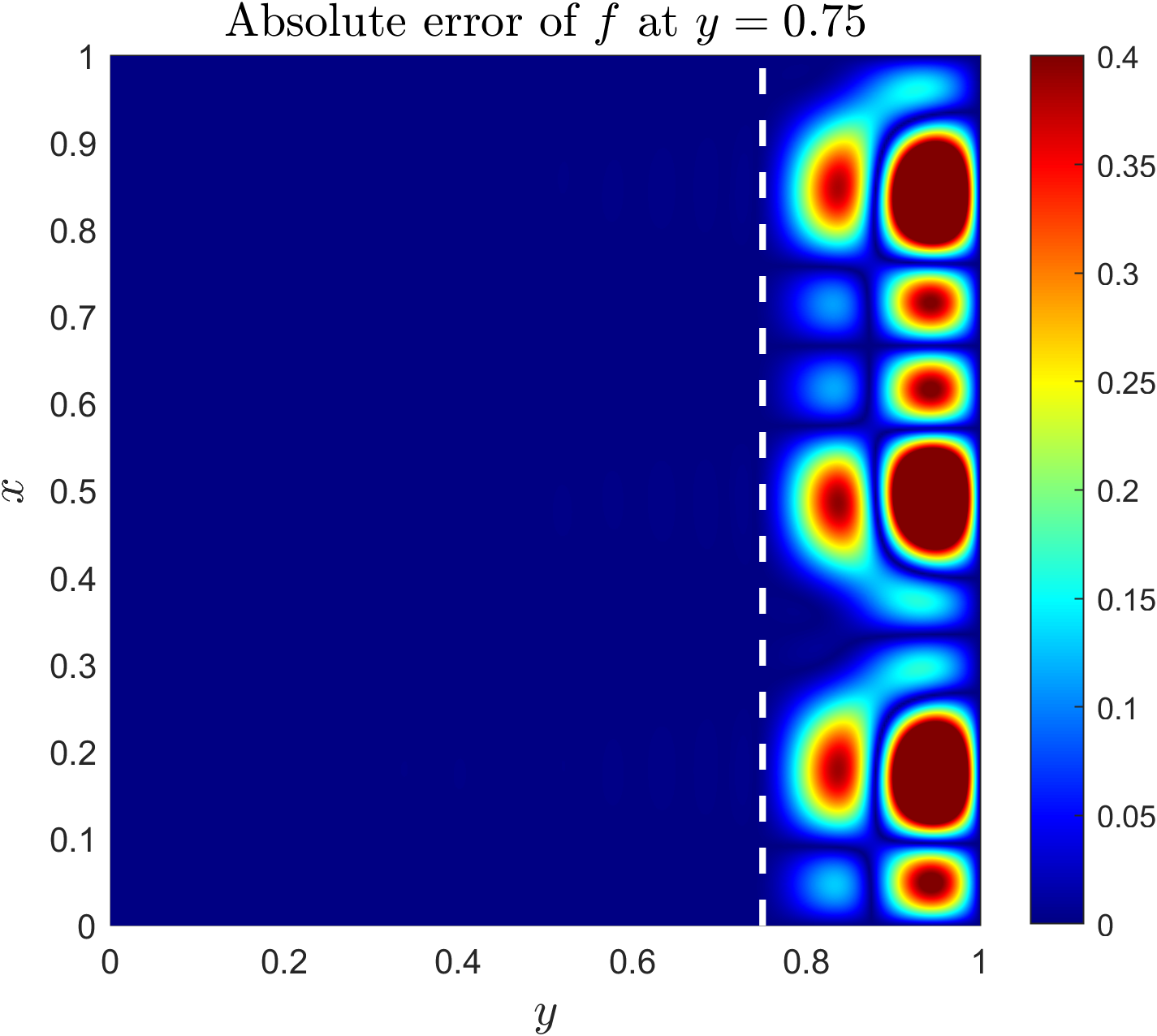}
        \includegraphics[scale=.3]{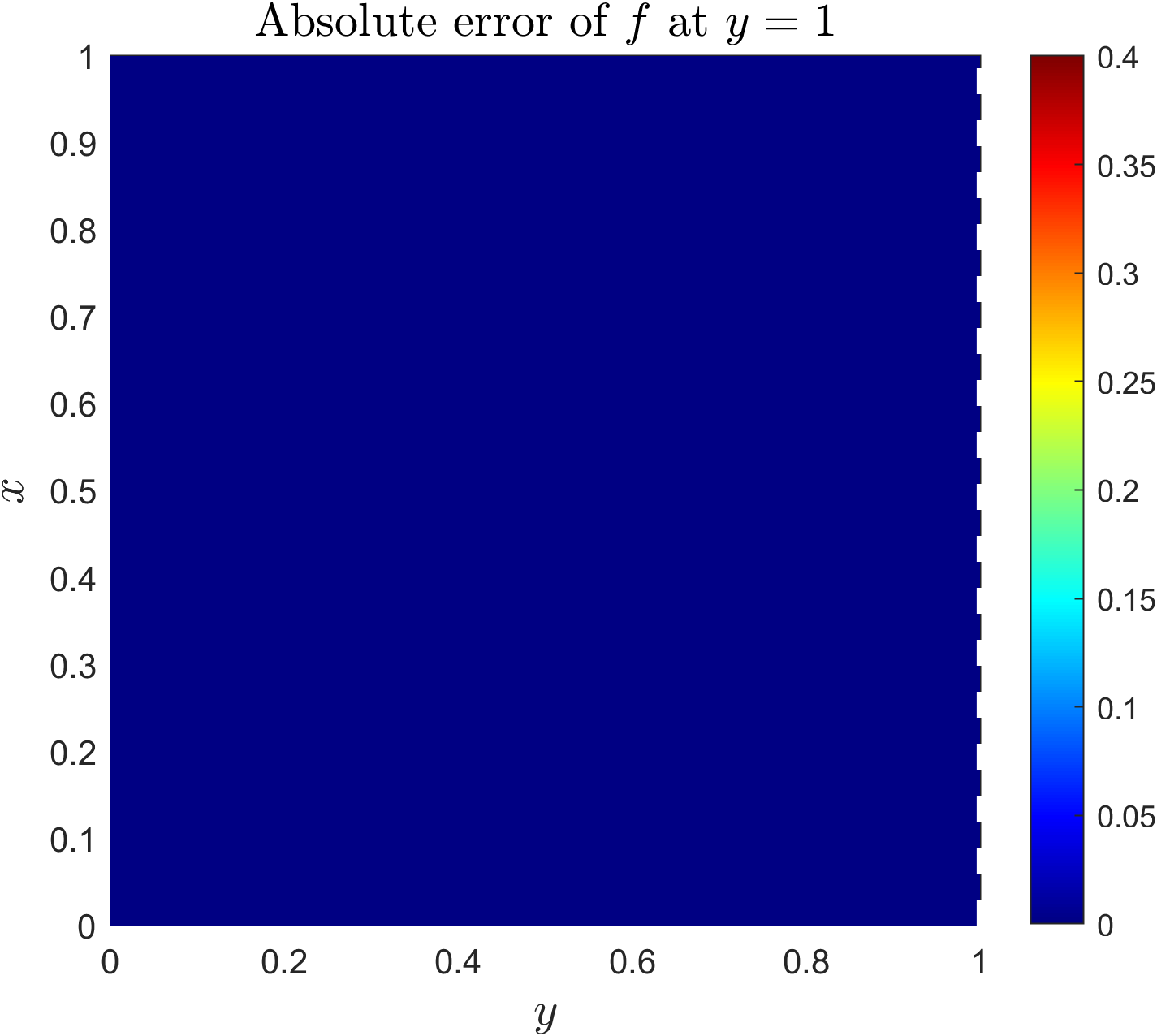}
    }
    \caption{Continual learning of the solution $u$ and source term $f$ of the 2D Poisson equation \eqref{eq:poisson} using our Riccati-based approach. White dashed lines: where integration has advanced in $y$ so far. % Quantitative results can be found in Table \ref{tab:example2}. 
    Information of $f$ is discretized in $x$ and then propagated along $y$. Hence, our approach only requires access to 1D slices of the domain instead of the entire domain, which highlights the potential memory benefits of our Riccati-based approach. }
    \label{fig:example2}
\end{figure}

%\update{change L to 1}
\noindent In this example, we extend our Riccati-based approach to a higher dimensional problem to demonstrate the potential memory advantages of our approach. Consider the 2D Poisson equation:
\begin{equation}\label{eq:poisson}
    \begin{dcases}
        -\kappa\Delta u(x, y) = f(x, y), &(x, y)\in (\Omega \setminus \partial \Omega),\\
        u(x, y) = 0, &(x, y)\in\partial\Omega,
    \end{dcases}
\end{equation}
where $\Omega:=[0, 1]^2$ is the domain, $\kappa = \frac{0.01}{\pi^2}$ is a constant, and $f:\Omega\to\R$ is the source term.
%where $\Omega:=[0, L]^2$ is the domain, $\kappa = \frac{0.01L}{\pi}$ is a constant, $L=1$, and $f$ is the source term derived from 
%\begin{equation}\nonumber
%    u(x, y) = -0.8 \sin(3\pi x)\sin(8\pi y) + 0.4 \sin(9\pi x) \sin(7\pi y) - 0.3 \sin(6\pi x)\sin(10\pi y).
%\end{equation}
Assume that we only have access to a few slices of $f$ due to limited computational resources, which prevents direct deployment of most traditional numerical solvers, which typically require full information of $f$ to solve the equation either iteratively or in one-shot. Hence, we can solve this PDE using continual learning to take advantage of the memory benefits of data streaming. Namely, we will avoid having to store the entire discretized grid of $\Omega$ by only processing/accessing information on a small portion of $\Omega$ at any given time. 
%In this regard, we transform this problem into a learning problem and employ our approach to solve the equation in a continual learning paradigm. 
We learn a linear model $(x,y)\mapsto\sum_{i=1}^n \theta_i\MLbasis_i(x, y)$ to approximate the solution $u$, where the coefficients $\weightvec = [\weight_1, \dots, \weight_n]^T$ are learned by minimizing:
%and establish the following loss function:
%\begin{equation}\label{eq:loss_pde}
%    \mathcal{L}(\weightvec) = \frac{\lambda_f}{2}\int_{[0, L]}\int_{[0, L]} |\kappa\Delta u_\weightvec(x, y) + f(x, y)|^2 dx dy + \frac{1}{2}\sum_{i=1}^n \gamma_i |\theta_{i}|^2,
%\end{equation}
%where $\lambda_f$ and $\gamma_i, i=1,...,n$ are belief weights. Note that we omit the boundary loss term in \eqref{eq:loss_pde} since the basis functions $\phi_i$ already satisfy the boundary conditions. 
%To match the framework in Section~\ref{sec:method}, we  discretize the integral-type loss~\eqref{eq:loss_pde} along the $x$-axis and treat the integration along the $y$-axis as the evolution in ``time'' of the corresponding HJ PDE. We use a uniform mesh of size $N$ on $[0, L]$ to approximate the integration $\int_{[0, L]} |\Delta u_\weightvec(x, y) - f(x, y)|^2 dx$ with the discrete sum $\sum_{i=1}^{N+1} |\Delta u_\weightvec(x_i, y) - f(x_i, y)|^2 h_x$, where $h_x = \frac{L}{N}$. Thus, the loss function~\eqref{eq:loss_pde} can be rewritten as:
\begin{equation}\label{eq:loss_pde_discrete}
    \mathcal{L}(\weightvec; t) = \frac{\lambda_f}{2N} \int_0^t \sum_{j=1}^{N+1}\left[\kappa\sum_{i=1}^n\weight_{i}\left(\frac{\partial^2\MLbasis_i}{\partial x^2} + \frac{\partial^2\MLbasis_i}{\partial y^2}\right)(x_j, y) + f(x_j, y)\right]^2dy + \frac{1}{2}\sum_{i=1}^n \gamma_i|\theta_i|^2,
\end{equation}
%\update{change $h_x$ to $1/N$, check matrix notation} 
where $\lambda_f$ and $\gamma_i, i=1,...,n$ are belief weights. Note that we omit the boundary loss in \eqref{eq:loss_pde_discrete} since the boundary conditions will be enforced by our choice of $\{\phi_i\}_{i=1}^n$. We also have discretized the integral-type loss in $x$ with a uniform mesh $x_j = \frac{j-1}{N}, j = 1, \dots, N+1$, 
%$\{x_j\}_{j=1}^{N+1}$ of size $N$ on $[0,1]$, 
so that the integration along $y$ can be treated as
the evolution in ``time'' of the corresponding HJ PDE. %We use a uniform mesh of size $N$ on $[0, L]$ to approximate the integration $\int_{[0, L]} |\Delta u_\weightvec(x, y) - f(x, y)|^2 dx$ with the discrete sum $\sum_{i=1}^{N+1} |\Delta u_\weightvec(x_i, y) - f(x_i, y)|^2 h_x$, where $h_x = \frac{L}{N}$. 
The above loss function~\eqref{eq:loss_pde_discrete} is in the same form of~\eqref{eq:loss_function} by setting $\lambda = \frac{\lambda_f}{N}$, $[\MLbasismat(\cdot)]_{ji} = \phi_i(x_j,\cdot)$, $[\mathcal{A}\MLbasismat(\cdot)]_{ji} =  \kappa(\frac{\partial^2\MLbasis_i}{\partial x^2} + \frac{\partial^2\MLbasis_i}{\partial y^2})(x_j, \cdot)$, $\MLy(\cdot)=-[f(x_1, \cdot), \dots, f(x_{N+1}, \cdot)]^T$, $\regmat = \Gamma^{1/2}$, where $\Gamma$ is a diagonal matrix whose $i$-th entry is $\gamma_i$, and $\regcenter = 0$. Thus, this learning problem~\eqref{eq:loss_pde_discrete} can be solved using the Riccati ODEs~\eqref{eqt:RiccatiODEs} % with initial conditions $\Sxx(0) = \Gamma^{-1}$, $\Sx(0) = \mathbf{0}$. 
to perform the integration along $y$, and instead of storing information of $f$ on all of $\Omega$, we only require information along 1D slices of $\Omega$ at any given time, which provides memory advantages over more traditional discretization methods. %In general, the $t$-direction (i.e., the direction along which information is propagated) should be chosen to be along a characteristic of the PDE.
%%% JD removed it, because it ``only" works when characteristics exist. Ellipctic equation has issue, and we considered in in the example. 

In our numerical experiments, we consider the case where $f$ is derived from 
\begin{equation}\nonumber
    u(x, y) = -0.8 \sin(3\pi x)\sin(8\pi y) + 0.4 \sin(9\pi x) \sin(7\pi y) - 0.3 \sin(6\pi x)\sin(10\pi y)
\end{equation}
(but $u$ is assumed to be unknown a priori), $N=400$, $\lambda=100$, $\gamma_i=1, i=1,...,n$, $\{\phi_i\}_{i=1}^{m^2} = \{(x,y)\mapsto\sin(j\pi x)\sin(k\pi y)\}_{j, k=1}^m$, $m=15$ ($n = m^2$), and we employ RK4 with step size $h=10^{-5}$. We also assume that our measurements of $f$ are corrupted with additive Gaussian noise with mean zero and scale $0.05$, which justifies the use of $\ell^2$-regularization. 
% In Figure \ref{fig:example2} and Table \ref{tab:example2}, we see that our Riccati-based approach obtains increasingly accurate inferences of $f$ and $u$ as more noisy data is incorporated into our learned models. In particular, the accuracy improves in both the regions we have already visited and future regions despite only accessing information along one 1D slice of the domain at a time. Thus, our methodology again avoids catastrophic forgetting by encoding all of the previous information in the solution to the corresponding HJ PDE.
%{\color{red}
In Figure \ref{fig:example2}, we see that our Riccati-based approach obtains increasingly accurate inferences of $f$ and $u$ as more noisy data is incorporated into our learned models. % and that the accuracy improves in both the regions we have already visited and future regions, indicating that we have successfully avoided catastrophic forgetting. 
Specifically, the relative $L_2$ errors of the inferences of $u$ are $81.42\%$, $34.00\%$, $0.07\%$ and of $f$ are $81.87\%$, $30.30\%$, $0.03\%$ once $y=0.25, 0.75, 1$ has been visited, respectively. 
%Similarly, the relative $L_2$ errors of the inferences of $f$ are $81.87\%$, $30.30\%$, $0.03\%$ once $y=0.25, 0.75, 1$ has been visited, respectively. 
%and $f$ are decreasing from $81.42\%$ and $81.87\%$ at $y=0.25$, to $34.00\%$ and $30.30\%$ at $y=0.75$, and $0.07\%$ and $0.03\%$ at $y=1$. 
For comparison, we also compute the LSE for~\eqref{eq:loss_pde_discrete}, in which the integral is approximated using Monte Carlo with $10^4$ uniform sampling points. The relative $L_2$ errors of the LSE inferences of $u$ and $f$ are $0.30\%$ and $0.09\%$, respectively. 
We compute all of the above $L_2$ errors using trapezoidal rule with a uniform $401\times 401$ grid over all of $\Omega$.
Note that LSE requires information of all of $\Omega$ at once, yet our approach obtains more accurate inferences once all of $\Omega$ has been visited despite only accessing information along one 1D slice of $\Omega$ at a time. The accuracy of the LSE inferences could be improved by increasing the number of Monte Carlo points. However, the LSE with $10^5$ Monte Carlo points is not able to be computed on a standard laptop (13th Gen Intel(R) Core(TM) i9-13900HX with 2.20 GHz processor and 16 GB RAM) due to memory constraints, which highlights the memory advantages of our approach. In this case, LSE would require an $N\times 10^5$ grid of $\Omega$, whereas our approach only requires $N$ points at a time, where the accuracy in $y$ can be improved without any additional memory burden by decreasing $h$.
%\update{trapezoidal rule with uniform grid of size $401\times 401$, $L_2$ error computed over all of $\Omega$}
%Thus, our methodology again avoids catastrophic forgetting by encoding all of the previous information in the solution to the corresponding HJ PDE.
%}

% \begin{table}[ht]
%     \footnotesize
%     \centering
%     \begin{tabular}{c|c|c|c|c}
%     \hline\hline
%        & $y=0.25$ & $y=0.75$ & $y=1$ & Reference \\
%        \hline
%        Error of $u$  & $81.42\%$ & $34.00\%$ & $0.07\%$ & $0.30\%$\\
%        \hline
%        Error of $f$  & $81.87\%$ & $30.30\%$ & $0.03\%$ & $0.09\%$ \\
%     \hline\hline
%     \end{tabular}
%     \caption{Relative $L_2$ errors of the inferences of the solution $u$ and source term $f$ of the PDE~\eqref{eq:poisson} at different $y$. %The metric is the relative $L_2$ error. 
%     Both errors improve as more information is incorporated into our learned models.
%     ``Reference'' is given by the LSE for \eqref{eq:loss_pde}, in which the integral-type loss is approximated using the Monte Carlo method with $10^4$ uniform sampling points. ({\color{blue}Comment from ZZ (to-be-deleted): the reference method could be improved by using more sampling points. However, my laptop ran out of memory when I used $10^5$ sampling points (this is a 2D problem).}) Note that LSE does not meet the requirements of continual learning as it requires access to the full information of $f$ at once.}
%     \label{tab:example2}
% \end{table}

\section{Summary}\label{sec:conclusion}
In this paper, we established a new theoretical connection between regularized learning problems with integral-type losses and the generalized Hopf formula for HJ PDEs with time-dependent Hamiltonians. This connection yields a new interpretation for certain SciML applications. Namely, when we solve these learning problems, we also solve an optimal control problem and its associated HJ PDE with time-dependent Hamiltonian. In the special case of linear regression, we leveraged this connection to develop a new Riccati-based methodology that provides promising computational and memory advantages in continual learning settings, while inherently avoiding catastrophic forgetting. 

This work opens opportunities for many exciting future directions. For example, in contrast to the methodology in~\cite{chen2023leveraging}, our methodology (Section~\ref{sec:riccati}) does not allow data to be added or removed in arbitrary order %nor does it allow the data fitting and regularization weights to be tuned without restarting the corresponding Riccati ODEs
since the corresponding HJ PDE must be evolved continuously in time. Achieving the same flexibility as the methodology in~\cite{chen2023leveraging} requires relaxing the convexity and regularity assumptions for the generalized Hopf formula in Section~\ref{sec:theory}. Thus, a worthwhile future direction of this work would be to extend to nonconvex and/or discontinuous Hamiltonians, which, in turn, would extend our connection to be between more general learning problems (e.g., those with nonconvex loss functions) and differential games (\cite{evans1984differentialgames}), instead of optimal control. 
In Section~\ref{sec:examples_pde}, we extended our Riccati-based approach to a 2D example, and as an illustration we propagated the information along the $y$-axis. For more general problems, it would be valuable to perform a more in-depth investigation into selecting an appropriate propagation direction, so that the memory load of this approach remains sufficiently reduced in higher dimensions.
%Another interesting future direction of this work would be to generalize our theory to viscous HJ PDEs, which have known connections with Bayesian modeling (\cite{darbon2021bayesian}) and thus would allow us to consider machine learning applications in Bayesian inference. 
Another natural extension would be to explore how our connection could be leveraged to reuse existing efficient machine learning algorithms to solve high-dimensional HJ PDEs and optimal control problems as so far, we have only explored the opposite direction. %However, in general, solving optimal control problems with time-dependent costs and dynamics is challenging and thus developing efficient algorithms for high-dimensional, time-variant optimal control problems would be transformative for the optimal control community.

%discuss adding/removing points in arbitrary order, hyperparameter tuning -- results in less regularity.. differential games.. order does not commute
%benefit is not storing the data (not necessarily continual learning -- often we access the source term as function instead of discrete measurements), the order we follow doesn't matter (but should follow the direction of the characteristics), arbitrarily choose a direction to propagate in (e.g., a dimension to treat as the time variable), we illustrate
%characteristic of the PDE, e.g. source term, boundary condition

%\newpage %acknowledgements and references do not count toward 10-page limit
\acks{P.C. is funded by the Office of Naval Research (ONR) In-House Laboratory Independent Research Program (ILIR) at NAWCWD managed by Dr. Claresta Dennis and a Department of Defense (DoD) SMART Scholarship for Service SEED Grant. (Approved for public release. Distribution is unlimited. PR 24-0095.)
Z.Z., J.D., and G.E.K. are supported by the DOE-MMICS SEA-CROGS project (DE-SC0023191) and the MURI/AFOSR project (FA9550-20-1-0358). G.E.K. is also supported by the ONR Vannevar Bush Faculty Fellowship (N00014-22-1-2795).}

\bibliography{references}

\end{document}